\documentclass[10pt,twocolumn,letterpaper]{article}

\usepackage{iccv}      

\def\iccvPaperID{5619} 

\usepackage[accsupp]{axessibility}  
\usepackage[hyphens]{url}
\usepackage{xcolor}
\usepackage[pagebackref=true,breaklinks=true,colorlinks,bookmarks=false]{hyperref}
\definecolor{deepred}{HTML}{940000}
\hypersetup{linkcolor=deepred}
\hypersetup{citecolor=[rgb]{0.4,0.15,0.95}}

\usepackage{colortbl}
\usepackage{graphicx}
\usepackage{amsmath}
\usepackage{amssymb}
\usepackage{booktabs}
\usepackage{multirow}
\usepackage{cuted}
\usepackage{bm}
\usepackage{xspace}
\usepackage{caption}
\usepackage{balance}
\usepackage{pifont}
\usepackage{times}
\usepackage[normalem]{ulem}
\usepackage{arydshln}
\usepackage{lipsum}
\usepackage{makecell}
\usepackage{bbding}

\usepackage{threeparttable}

\usepackage{setspace}
\usepackage{enumitem}

\usepackage[toc,page]{appendix}

\definecolor{Gray}{gray}{0.92}
\definecolor{ForestGreen}{RGB}{34,139,34}
\definecolor{Cyan}{rgb}{0.0, 0.72, 0.92}
\definecolor{YellowOrange}{RGB}{255,174,66}

\newcolumntype{a}{>{\columncolor{Gray}}c}

\newlength\savewidth\newcommand\shline{\noalign{\global\savewidth\arrayrulewidth
  \global\arrayrulewidth 1pt}\hline\noalign{\global\arrayrulewidth\savewidth}}

\usepackage{float}

\definecolor{Bittersweet}{rgb}{0.0,0.0,0.0}

\newcommand{\supmatCOLOR}{Bittersweet}

\newcommand{\ourrefcolor}{\color{black}}

\newcommand{\qheading}[1]{\noindent\textbf{#1}:}

\newcommand{\nospaceheading}[1]{\noindent\textbf{#1}}

\newcommand{\colorRef}[1]{\textcolor{black}{#1}} 
\usepackage[capitalize]{cleveref}
\crefname{figure}{\colorRef{Fig.}}{\colorRef{Figs.}}
\Crefname{figure}{\colorRef{Figure}}{\colorRef{Figures}}
\crefname{section}{\colorRef{Sec.}}{\colorRef{Secs.}}
\Crefname{section}{\colorRef{Section}}{\colorRef{Sections}}
\Crefname{table}{\colorRef{Table}}{\colorRef{Tables}}
\crefname{table}{\colorRef{Tab.}}{\colorRef{Tabs.}}

\newcommand{\Fig}[1]{\mbox{\ourrefcolor{{Figure}}~\ref{#1}}}

\newcommand{\tab}[1]{\mbox{\ourrefcolor{{Table}}~\ref{#1}}}
\newcommand{\fig}[1]{\mbox{\ourrefcolor{{Figure}}~\ref{#1}}}

\newcommand{\methodname}{{ELICIT}\xspace}
\newcommand{\mn}{\methodname}

\newcommand{\ourTitle}{\vspace{-1mm}One-shot Implicit Animatable Avatars with Model-based Priors}

\newcommand{\cmark}{\small\ding{51}}
\newcommand{\xmark}{\small\ding{55}}

\newcommand{\smpl}{\mbox{SMPL}\xspace}

\newcommand{\normabs}[1]{\left\lVert#1\right\rVert_1}
\newcommand{\normmse}[1]{\left\lVert#1\right\rVert_2^2}

\DeclareSymbolFont{matha}{OML}{txmi}{m}{it}
\DeclareMathSymbol{\varv}{\mathord}{matha}{118}

\newcommand{\supmat}{{\mbox{\color{\supmatCOLOR}{Sup.~Mat.}}}\xspace}

\renewcommand{\ie}{\mbox{i.e.}\xspace}
\renewcommand{\eg}{\mbox{e.g.}\xspace}

\iccvfinalcopy
\begin{document}

\title{\ourTitle}

\author{
    Yangyi Huang\textsuperscript{1,4*}\quad
    Hongwei Yi\textsuperscript{2*}\quad
    Weiyang Liu\textsuperscript{2,3}\quad
    Haofan Wang\textsuperscript{4} \\
    Boxi Wu\textsuperscript{5}\quad
    Wenxiao Wang\textsuperscript{5}\quad
    Binbin Lin\textsuperscript{5,6\dag}\quad
    Debing Zhang\textsuperscript{4}\quad
    Deng Cai\textsuperscript{1}
    \\[1mm]
    \textsuperscript{1} State Key Lab of CAD \& CG, Zhejiang University
    \\
    \textsuperscript{2} Max Planck Institute for Intelligent Systems, T{\"u}bingen
    \quad
    \textsuperscript{3} University of Cambridge
    \\
    \textsuperscript{4} Xiaohongshu Inc.
    \quad
    \textsuperscript{5} School of Software Technology, Zhejiang University
    \quad
    \textsuperscript{6} Fullong Inc.
    \\
    {\tt\small huangyangyi@zju.edu.cn
    \quad
    hongwei.yi@tuebingen.mpg.de
    }
     \\
    \vspace{-0.7em}
}

\newcommand{\teaserCaption}{
Our method creates free-viewpoint motion videos from a single image by constructing an animatable NeRF representation in one-shot learning.}

\twocolumn[{
    \renewcommand\twocolumn[1][]{#1}
    \maketitle
    \centering
    \vspace{-6mm}
    \begin{minipage}{1.0\textwidth}
        \centering
        \includegraphics[width=1.0 \linewidth]{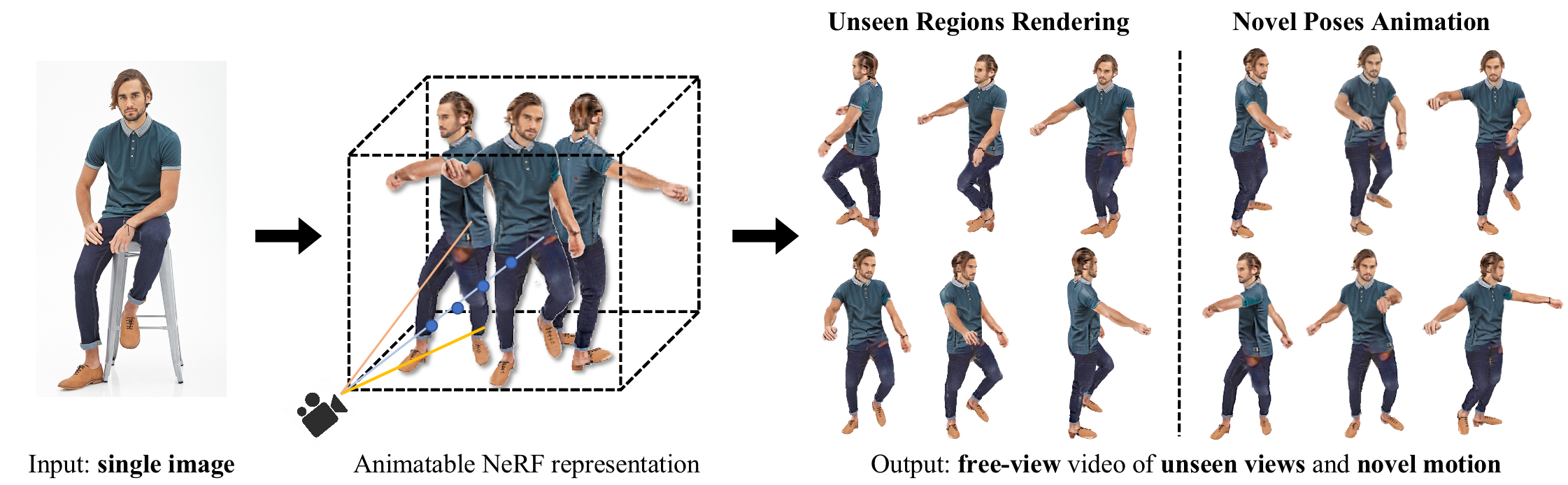}
    \end{minipage}
    \vspace{-2mm}
    \captionof{figure}{\footnotesize\teaserCaption }
    \vspace{6mm}
    \label{fig:teaser}
}]

\def\thefootnote{*}\footnotetext{Equal contribution.}
\def\thefootnote{\dag}\footnotetext{Corresponding author.}

\vspace{1.5mm}
\begin{abstract}
\vspace{-2mm}
Existing neural rendering methods for creating human avatars typically either require dense input signals such as video or multi-view images, or leverage a learned prior from large-scale specific 3D human datasets such that reconstruction can be performed with sparse-view inputs. Most of these methods fail to achieve realistic reconstruction when only a single image is available. To enable the data-efficient creation of realistic animatable 3D humans, we propose \mn, a novel method for learning human-specific neural radiance fields from a single image. Inspired by the fact that humans can effortlessly estimate the body geometry and imagine full-body clothing from a single image, we leverage two priors in \mn: 3D geometry prior and visual semantic prior. Specifically, \mn utilizes the 3D body shape geometry prior from a skinned vertex-based template model (i.e., SMPL) and implements the visual clothing semantic prior with the CLIP-based pre-trained models. Both priors are used to jointly guide the optimization for creating plausible content in the invisible areas. Taking advantage of the CLIP models, \mn can use text descriptions to generate text-conditioned unseen regions. In order to further improve visual details, we propose a segmentation-based sampling strategy that locally refines different parts of the avatar. Comprehensive evaluations on multiple popular benchmarks, including ZJU-MoCAP, Human3.6M, and DeepFashion, show that \mn outperforms strong baseline methods of avatar creation when only a single image is available.
\ificcvfinal The code is public for research purposes at \url{https://huangyangyi.github.io/ELICIT} \fi 
\end{abstract}

\vspace{-3mm}
\section{Introduction} \label{sec:introduction}

Creating realistic 3D contents of animatable human avatars from readily available camera inputs is of great significance for AR/VR applications, such as telepresence, virtual fitness, and so on.
It is quite a challenging task and requires disentangled reconstruction of 3D geometry, the appearance of a clothed human, and accurate modeling of complex body poses for animation.

Current human-specific neural rendering methods have achieved promising performance when dense and well-controlled inputs are available, \eg, multi-view videos captured by well-calibrated multi-camera systems~\cite{pengNeuralBodyImplicit2021, xuHNeRFNeuralRadiance2021, wuMultiViewNeuralHuman2020, pengAnimatableNeuralRadiance2021a, zhangEditableFreeviewpointVideo2021a}, or long monocular videos~\cite{wengHumanNeRFFreeViewpointRendering2022} where almost all parts of the human body are visible. Despite their excellent performance, it is inconvenient (sometimes impossible) for ordinary users to obtain such high-quality dense inputs. Various methods have been proposed to address this data inefficiency. For example, ARCH~\cite{huangARCHAnimatableReconstruction2020} and ARCH++~\cite{heARCHAnimationReadyClothed2021} train reconstruction models with a single image input on large 3D scans datasets, but they do not generalize well to in-the-wild data. Neural radiance fields~(NeRF)~\cite{mildenhall2021nerf} based human-specific methods~\cite{gaoMPSNeRFGeneralizable3D2022, liuNeuralActorNeural2021a, kwonNeuralHumanPerformer2021} train conditional models on multi-view images or video datasets to improve generalizability. However, when only sparse-view inputs are available, they also fail to generate realistic results under extreme settings, \eg, single monocular images.

Instead of learning conditional models from large-scale datasets~\cite{yuPixelNeRFNeuralRadiance2021a, chenMVSNeRFFastGeneralizable2021}, recent work introduces various regularizations for geometry~\cite{niemeyerRegNeRFRegularizingNeural2022} and appearance~\cite{jainPuttingNeRFDiet2021, xuSinNeRFTrainingNeural2022a} to avoid degeneration, which makes it possible to synthesize visually plausible views in a semi-supervised framework without extra training data. 

However, due to the missing information about the occluded areas of the subject, they can hardly synthesize unseen views that barely overlap with the input views. To address these limitations, we propose a novel method, \mn, to learn human-specific neural radiance fields from a single image. We explicitly take advantage of the body shape geometry prior and the visual clothing semantic prior to guide the optimization and achieve free-view rendering from single images.

In summary, our contributions are listed below:

\vspace{0.7mm}
\begin{itemize}[leftmargin=*,nosep]
\setlength\itemsep{0.3em}
    \item We present \mn, a novel approach that can train an animatable neural radiance field from a single image without relying on extra training data.

    \item We propose two effective model-based priors to achieve an animatable 3D free-view rendering digital avatar from single image: 1) the visual clothing semantic prior. Specifically, we leverage the power of large pretrained vision-language models (\ie, CLIP) to hallucinate the unseen parts of the clothed body.
2) the human shape prior from the SMPL model. 
We use the estimated SMPL body shape and pose to constrain our reconstructed clothed 3D avatar be consistent with it.

    \item To create more realistic and consistent body part details, we propose a novel sampling strategy conditioned on the SMPL semantic segmentation and body rotation.
\end{itemize}

We conduct both quantitative and qualitative comparisons with recent human-specific neural rendering methods in the setting of single image input. We observe that ELICIT can consistently outperform existing methods in both free-view rendering and avatar animation, and simultaneously demonstrate promising performance on in-the-wild images.

\section{Related Work} \label{sec:relatedwork}
\vspace{-.5mm}
\begin{table}[t]
\centering
\scriptsize
\setlength{\tabcolsep}{2.5pt}
\vspace{0.5mm}
\begin{tabular}{c|cccc}
Method &
  \multicolumn{1}{c}{Subject data} &
  \multicolumn{1}{c}{\begin{tabular}{c}Extra\\training data\end{tabular}} &
  \multicolumn{1}{c}{\begin{tabular}[c]{@{}c@{}}Invisible area \\ completion\end{tabular}} &
  \multicolumn{1}{c}{Animatable} \\ \shline
\begin{tabular}{c}
NeuralBody~\cite{pengNeuralBodyImplicit2021}\\
Ani-NeRF~\cite{pengAnimatableNeuralRadiance2021a}\\
HumanNeRF~\cite{wengHumanNeRFFreeViewpointRendering2022}\\
\end{tabular}
 &
  \begin{tabular}{c}
  multi-view\\
  images,\\
  monocular\\
  videos\end{tabular} &
  \textbf{data-free} &
  \xmark &
  \cmark \\ \hline
\begin{tabular}{c}
PiFU~\cite{saitoPIFuPixelAlignedImplicit2019} \\ PaMIR~\cite{zheng2021pamir} \\ ARCH~\cite{huangARCHAnimatableReconstruction2020}\\
ARCH++~\cite{heARCHAnimationReadyClothed2021} \\
PHORHUM~\cite{alldieck2022photorealistic}
\end{tabular} & \begin{tabular}{c}\textbf{monocular}\\\textbf{images}\end{tabular} & 3D scans          & \cmark & \cmark \\ \hline
\begin{tabular}{c}
MPS-NeRF~\cite{gaoMPSNeRFGeneralizable3D2022} \\ NHP~\cite{kwonNeuralHumanPerformer2021}
\end{tabular} &
  \begin{tabular}{c}sparse videos,\\
  multi-view\\
  images
  \end{tabular} &
  \begin{tabular}{c}multi-view\\videos\end{tabular} &
  \xmark &
 \cmark \\\hline
MonoNHR~\cite{choiMonoNHRMonocularNeural2022}                                                             & \begin{tabular}{c}\textbf{monocular}\\\textbf{images}\end{tabular} & \begin{tabular}{c}multi-view\\images\end{tabular} & \cmark & \xmark    \\\hline
EVA3D~\cite{hongEVA3DCompositional3D2022}                                                             & \begin{tabular}{c}\textbf{monocular}\\\textbf{images}\end{tabular} & \begin{tabular}{c}monocular\\images\end{tabular} & \cmark & \cmark    \\\hline
\rowcolor{Gray}
\mn~(ours) &
  \begin{tabular}{c}\textbf{monocular}\\\textbf{images}\end{tabular} &
  {\textbf{data-free}} &
  {\cmark} &
  {\cmark} \\ 
\end{tabular}
\vspace{-1.5mm}
\caption{\footnotesize\label{table:methodComparison} \textbf{Recent human rendering methods that are most relevant to our work.} \mn is the first work that satisfies these four characteristics together: 1) only requires a single monocular image as an input. 2) doesn't need extra training data of the subject person. 3) supports recovering body areas that are invisible from the given input view. 4) animatable.}
\vspace{-3mm}
\end{table}

\nospaceheading{Animatable human neural rendering.} Existing methods of animatable human-specific neural rendering can be divided into 2D-based methods and 3D-based methods. 2D-based methods are mostly derived from image-based human pose transfer methods~\cite{sarkarStylePoseControl2021, maPoseGuidedPerson2017, neverovaDensePoseTransfer2018, balakrishnanSynthesizingImagesHumans2018}, leveraging explicit temporal constraints~\cite{chanEverybodyDanceNow2019, yangPoseGuidedHuman2018}, optical flow estimation~\cite{wangVideotoVideoSynthesis2018a}, and warping field~\cite{siarohinFirstOrderMotion2019, yoonPoseGuidedHumanAnimation2021} to create temporally consistent pose-guided videos from input videos or images. Most single-image-based 3D methods~\cite{saitoPIFuPixelAlignedImplicit2019, saito2020pifuhd, natsume2019siclope, li20193d} learn encoder-decoder models from high-quality human 3D scans data. 
Among these works, ARCH~\cite{huangARCHAnimatableReconstruction2020}, ARCH++~\cite{heARCHAnimationReadyClothed2021}, and PHORHUM~\cite{alldieck2022photorealistic} are some of the most promising methods for reconstructing animation-ready 3D representations. However, data-driven methods are limited by the diversity of their training data distribution and may struggle with generalization issues when dealing with unseen clothing styles and complex body poses.

Recent works about human-specific neural radiance fields reconstruct animatable 3D human NeRF representation from multi-view or single-view video (For NeRF, see \cite{mildenhall2021nerf}). Most of them do per-subject optimization on an implicit model, using the whole video sequence as training data. Among which  \cite{pengNeuralBodyImplicit2021} learns structured latent codes on SMPL~\cite{SMPL:2015} mesh vertices, other methods construct the representation in a canonical space by modeling pose-driven deformation~\cite{wengHumanNeRFFreeViewpointRendering2022,xuHNeRFNeuralRadiance2021, suANeRFArticulatedNeural2021, zhaoHumanNeRFEfficientlyGenerated2022, pengAnimatableImplicitNeural2022, pengAnimatableNeuralRadiance2021a}. While these methods produce impressive results, they require dense inputs that cover most areas of the human body. In contrast, our approach can generate an animatable realistic character from a single image, making it more user-friendly and flexible for a wider range of applications.

\nospaceheading{Single-view-based NeRF.} The setting of novel view synthesis from only a single image is challenging for NeRF-based methods because incomplete geometric information can lead to degeneration results. Also, it is difficult for the model to synthesize regions in the novel view which is not visible for the input due to occlusion. Some existing methods utilize learned prior about scene geometry and appearance in a data-driven manner, e.g., generative adversarial models~\cite{schwarzGRAFGenerativeRadiance2020,trevithickGRFLearningGeneral2021a}, supervised learning~\cite{yuPixelNeRFNeuralRadiance2021a,chenMVSNeRFFastGeneralizable2021, johariGeoNeRFGeneralizingNeRF2022, wangIBRNetLearningMultiView2021a}, and unsupervised learning~\cite{miIm2nerfImageNeural2022} for conditional NeRF. However, most of these methods only focus on simple 3D shapes~\cite{changShapeNetInformationRich3D2015}. Eg3D~\cite{chanEfficientGeometryAware3D2022} and CG-Nerf~\cite{joCGNeRFConditionalGenerative2021} are two representative methods that work on specific types of objects, such as human faces, using conditional generative NeRF.

There are also non-data-driven methods introducing priors from off-the-shelf models, including depth cues~\cite{dengDepthsupervisedNeRFFewer2022, liMINEContinuousDepth2021a} and other knowledge such as object geometry~\cite{niemeyerRegNeRFRegularizingNeural2022}.
SinNeRF~\cite{xuSinNeRFTrainingNeural2022a} and DietNeRF~\cite{jainPuttingNeRFDiet2021} use pre-trained image encoders to introduce semantic prior and produce semantically consistent novel view synthesis results from sparse inputs. Similarly, our work utilizes an SMPL-based human body prior and a CLIP-based visual semantic prior available in the task setting of single image-based human rendering and generates photo-realistic free-view renderings.

\nospaceheading{CLIP-driven radiance fields.}
CLIP~\cite{radfordLearningTransferableVisual2021b} is a cross-modality representation learning method that has recently been applied to text-driven image generation~\cite{rameshZeroShotTexttoImageGeneration2021, saharia2022photorealistic, ramesh2022hierarchical}. Several works have incorporated CLIP and radiance fields for 3D-aware synthesis tasks. DietCLIP~\cite{jainPuttingNeRFDiet2021} synthesizes view-consistent novel views from sparse view input with a CLIP-based loss as a regularization on NeRF. CLIP-NeRF~\cite{wangCLIPNeRFTextandImageDriven2021} applies joint image-text latent space in a conditioned NeRF for manipulation with multi-model inputs. LaTeRF~\cite{mirzaeiLaTeRFLabelText2022} uses CLIP loss to extract objects of interest from the scene, similar to texture cues. AvatarCLIP~\cite{hongAvatarCLIPZeroShotTextDriven2022} and Dream Fields~\cite{jain2022zero} apply CLIP to the optimization process for text-driven 3D generation. NeuralLift-360~\cite{xu2022neurallift} enables lifting a 3D object from a single image based on CLIP-based image similarity. In our work, we extend the use of CLIP-driven NeRF by leveraging it for human-specific rendering from a single image, exploring its potential in generating photo-realistic free-view renderings. 

\nospaceheading{Most relevant works.}
Recently, there have been several related works in the field of single-image-based human rendering. MonoNHR~\cite{choiMonoNHRMonocularNeural2022} proposes a data-driven approach using a conditional NeRF to render free-viewpoint images of a character from a single image input. EVA3D~\cite{hongEVA3DCompositional3D2022}, on the other hand, learns an unconditional 3D human generative model on the DeepFashion dataset~\cite{liuLQWTcvpr16DeepFashion} and can reconstruct 3D humans from a single image by GAN inversion~\cite{roich2021pivotal, corona2021smplicit}. However, its generalizability is largely limited by the biased distribution of the training datasets. A comparison of our method and related works is summarized in Table~\ref{table:methodComparison}. \mn only requires a single image as input without using extra training data, and yet supports both invisible area completion and body animation.

\section{Method} \label{sec:method}
\newcommand{\methodCaption}{
\textbf{Method overview.} Our method generates an animatable avatar from a single source image of a person, which can be used to create pose-guided free-view renderings of the person with any target motion in SMPL format. \mn train an animatable implicit human representation called HumanNeRF using one-shot prior-based learning. We use two model-based priors to guide the optimization process: the SMPL-based Geometric Prior and the Visual-Model-based Semantic Prior. The Human Body Prior is (a) initialized with multi-view video frames rendered by SMPL meshes and (b) uses a silhouette loss to constrain synthesized geometry and body poses during training. The Semantic Prior provides (c) pose-view-consistent semantic supervision for novel views of novel poses using a powerful pre-trained visual model. Additionally, we propose a Hybrid Sampling Strategy that includes (d) body-part-aware sampling to refine body-part details and (e) rotation-aware sampling to better recover heavily occluded views.

}

\begin{figure*}[htbp]
    \vspace{-5mm}
    \centering
    \includegraphics[width=1.0 \linewidth]{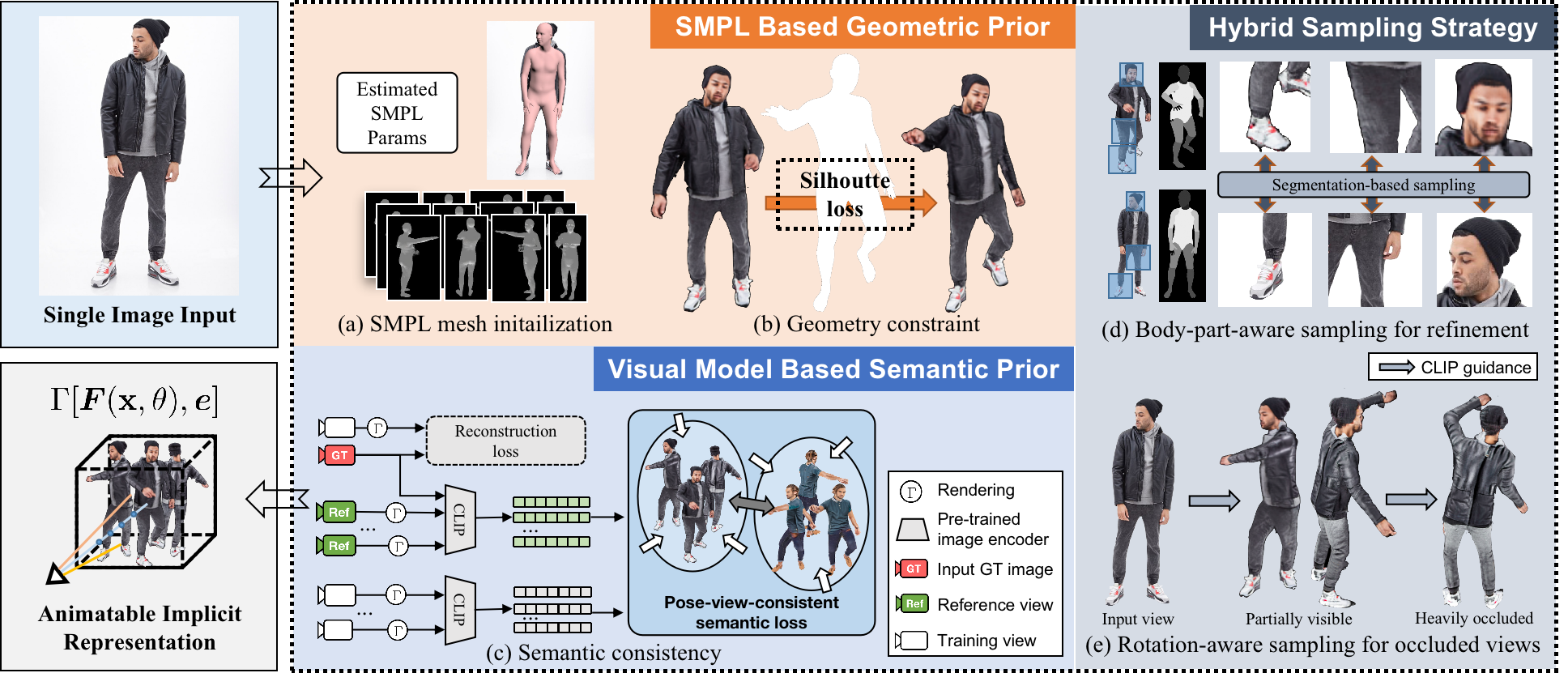}
    \vspace{-5mm}
    \captionof{figure}{\footnotesize\methodCaption}
    \label{fig:method}
    \vspace{-1mm}
\end{figure*}

\subsection{Problem Specification}
We formulate the task of creating free-view videos for a character in novel poses as follows. The input includes a single-view image $I_{s}$ of the character with camera parameters $\mathbf{e}_s$, SMPL-parameters $(\beta, \theta_s)$, where $\beta$ describes the body shape of the character, and $\theta_{s}$ describes the body pose of the character in the input image. We also input a motion sequence of length $n$ by SMPL pose parameters $\Theta_t = \{\theta_t^i\}_{i=1}^{L}$ and camera parameters of each frame $\boldsymbol E_t = \{\mathbf{e}_t^i\}_{i=1}^{L}$ for animation. The output is $n$ video frames $\{I_{t}^i\}_{i=1}^L$ rendered by pose-conditioned NeRF model under the given camera parameter, 
\begin{equation}
 I_t^i=\Gamma[\boldsymbol F(\mathbf{x}, \mathbf{\theta}_t^i), \mathbf{e}_t^i],
\end{equation}
where $\boldsymbol F$ is the pose-conditioned radiance field function and $\Gamma$ represents volume rendering.

\subsection{Preliminaries}

\textit{\smpl}~\cite{SMPL:2015}, Skinned Multi-Person Linear model, is a skinned vertex-based template model driven by large-scale aligned human surface scans. SMPL encodes posed body shape by a pose parameter $\theta_t^i\in \mathbb{R}^{72}$ and a shape parameter $\beta \in \mathbb{R}^{10}$, and outputs a blend shape sculpting the human body with 6890 vertices. We use SMPL parameters to represent the input character's body shape, posture, and pose sequence input of the target motion.

\textit{HumanNeRF}~\cite{wengHumanNeRFFreeViewpointRendering2022} is a human-specific variant of neural radiance field(NeRF), which supports free-view rendering of a moving character from monocular video inputs. In particular, HumanNeRF represents a moving character with a canonical appearance volume $F_c$ warped to an observed pose to produce output appearance volume $F_o$:
\begin{align}
    \vspace{-4mm}
F_o(\boldsymbol x, \boldsymbol p) = F_c(\mathcal{T}(\boldsymbol x, \boldsymbol p)),
    \vspace{-4mm}
\end{align}
where $F_c:\boldsymbol\rightarrow(\boldsymbol c, \sigma)$maps position $\boldsymbol x$ to color $\boldsymbol c$ and volume density $\sigma$. Notice that HumanNeRF uses a simplified version of NeRF without considering viewing directions. The motion field $\mathcal{T}:(\boldsymbol x_o, \boldsymbol p)\rightarrow \boldsymbol x_c$ maps positions in observed space back to the canonical space, conditioned by pose parameters $\boldsymbol p=(J, \Omega)$, where $J$ represents 3D joint locations and $\Omega$ represents local joint rotations. The novel views are synthesized by NeRF-based volume rendering:
\begin{align}
    \vspace{-4mm}
\boldsymbol C(\boldsymbol r)=\int_{t_n}^{t_f}T(t)\sigma(\boldsymbol r(t))\boldsymbol c(\boldsymbol r(t), \boldsymbol d)dt,
    \vspace{-4mm}
\end{align}
where the $T(t)$ is the transmittance of the light at position $t$, $T(t)=\exp (-\int_{t_n}^t\sigma(\boldsymbol r(s))ds)$. And $r$ is the pixel ray cast from the observer ,$\boldsymbol r(t)=\boldsymbol o + t\boldsymbol d$.
The original data-driven optimization of HumanNeRF requires monocular video input where most of the regions of the character are visible. We use it as the basic model of implicit neural representation for free-view motion rendering.

\subsection{Prior-driven One-shot Learning for Single-image Human Rendering}

\Fig{fig:method} illustrates our overall pipeline. \mn obtains the animatable implicit human representation by per-subject optimization with a single image input. We formulate this one-shot learning process as follows:

For each iteration, a training view with a character pose and camera parameters, $V_\textrm{train}=(\theta_\textrm{train}, \mathbf{e}_\textrm{train})$, is sampled from the input view  $V_s=(\theta_s, \mathbf{e}_s)$ and target views $V_t\in\{(\theta_i, \mathbf{e}_j)\}_{i=1,j=1}^{L,M}$, where $\{\mathbf{e}_j\}_{j=1}^M$ are  preset cameras around the character. We supervise the training view rendering with a respective reference view $V_\textrm{ref}$. The reference view could be the ground-truth view or rendered results of a sampled neighboring view.

On the one hand, to get realistic synthesis, rendering a consistent input view is the fundamental goal to be guaranteed. When the sampled view $V_\textrm{train}$ is identical to $V_s$, we select $V_\textrm{ref}=V_s$ and use the input image $I_s$ as the training target of rendered $\hat{I}_s$. We formulate our reconstruction loss the same as HumanNeRF~\cite{wengHumanNeRFFreeViewpointRendering2022}.
\begin{align}
    \vspace{-4mm}
    \mathcal{L_{\textrm{recon}}}=\mathcal{L}_{\textrm{LPIPS}}(I_s, \hat{I}_s) +\lambda\mathcal{L}_{\textrm{MSE}}(I_s, \hat{I}_s),
    \vspace{-4mm}
    \label{eq:recon}
\end{align}
where $\mathcal{L}_{\textrm{MSE}}$ is a pixel-wise mean square error loss, and $\mathcal{L}_{\textrm{LPIPS}}$ is a VGG-based perception loss that is robust to slight misalignment and improves reconstruction details.

On the other hand, we need to supervise $V_t$ for novel view synthesis and pose synthesis. We expect the synthesis results to have: (1) a consistent appearance with the input character, (2) a plausible geometry that approximates the actual clothed body shape, and (3) a body pose that matches the target motion. Obtaining such 3D-aware synthesis from incomplete input requires utilizing prior knowledge. In contrast to using a learned prior from multi-view images~\cite{gaoMPSNeRFGeneralizable3D2022, kwonNeuralHumanPerformer2021, choiMonoNHRMonocularNeural2022} or 3D scans training data~\cite{xiuICONImplicitClothed2022, saito2020pifuhd, saitoPIFuPixelAlignedImplicit2019,heARCHAnimationReadyClothed2021}, we introduce two model-based prior to guide the optimization. One is \textit{visual model-based semantic prior}, which supervises the synthesis of consistent visual contents. The other is \textit{SMPL-based human-specific prior} that provides 
knowledge about human body shape and posture.

\subsubsection{Visual model-based semantic prior}

Recent works~\cite{jainPuttingNeRFDiet2021, jain2022zero, xu2022neurallift} show that novel view synthesis from a single image or sparse inputs can be done with the guidance of an embedding loss, which enforces semantic consistency between unseen views and the reference view. Such optimization-based methods are also applicable to our task of synthesizing 3D-aware content for a clothed human. To achieve this, we need a powerful vision model to embed the images from different views of 3D humans in a semantically meaningful latent space. 

Among different models, we find that the CLIP~\cite{radfordLearningTransferableVisual2021b} visual encoders pre-trained on diverse image-text pairs data are suitable for this task. In \fig{fig:view-pose-consistent}, we carry out a similar evaluation of CLIP-NeRF~\cite{wangCLIPNeRFTextandImageDriven2021}, demonstrating view-pose-consistency of CLIP embeddings on human images. On the other hand, the CLIP models can also capture detailed visual semantics such that rich supervision signals can be utilized for a vivid generation~\cite{goh2021multimodal}.

\newcommand{\viewCaption}{
\textbf{View-pose-consistency of the CLIP embeddings.} The embedding distance of the same character under different views and poses is significantly smaller than the distance between two different characters.
}

\begin{figure}[htbp]
    \vspace{-1.5mm}
    \centering
    \includegraphics[width=1.00 \linewidth]{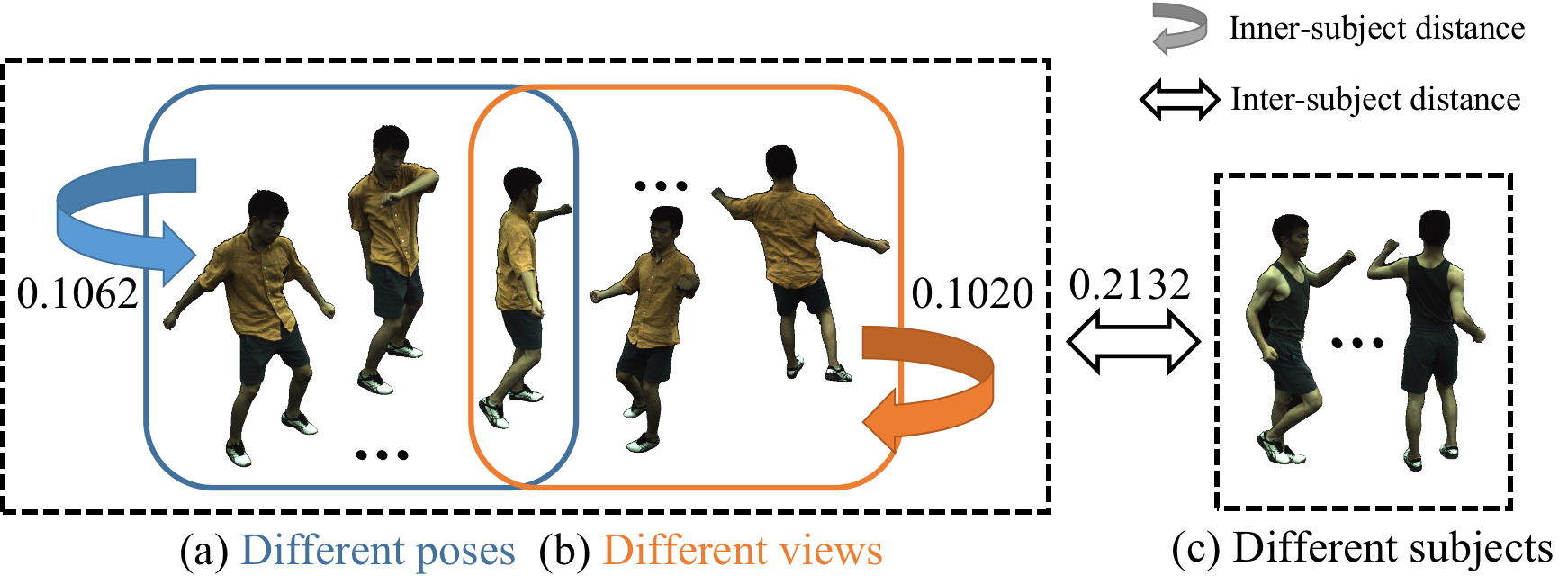}
    \vspace{-6mm}
    \captionof{figure}{\footnotesize\viewCaption}
    \label{fig:view-pose-consistent}
    \vspace{-0.5mm}
\end{figure}

 By comparing the performance of different model-based embedding losses, we select the CLIP ViT-based cosine distance as the semantic loss, formulated as follows:

\begin{align}
    \vspace{-4mm}
    \mathcal{L}_{\textrm{CLIP}}=\phi(I_\textrm{ref})^T\phi(\hat{I}_\textrm{train})
    \vspace{-4mm}, 
\end{align}where $\hat{I}_\textrm{train}$ is the rendered image of sampled training view, $I_\textrm{ref}$ is the reference view, and $\phi$ is the normalized embedding function of the CLIP ViT. 

Notably, the joint embedding space of CLIP is widely applied in the latest text-driven image generation works~\cite{wangCLIPNeRFTextandImageDriven2021, rameshZeroShotTexttoImageGeneration2021, fransCLIPDrawExploringTexttoDrawing2021, hongAvatarCLIPZeroShotTextDriven2022}. It also enables our method to support the use of user prompts $P_{\textrm{ref}}$ to guide the optimization of novel views. We can use the CLIP text embedding $\phi_{\textrm{text}}(P_{\textrm{ref}})$ as a reference in $\mathcal{L}_{\textrm{CLIP}}$. \Fig{fig:text-guided} demonstrates that detailed text prompts can aid in synthesizing invisible garments, but using only text guidance is insufficient for preserving the identity of the avatar. On the other hand, image-based semantic loss can recover crucial visual attributes like facial appearance and texture details. By utilizing both semantic losses, we can enhance the performance in challenging cases, such as those involving complex garments.
\newcommand{\textGuidedCaption}{
\textbf{Enhancing semantic prior with text prompts.} Combining text guidance with image guidance in the semantic loss helps recover the garment structure of the avatar's backside. It is worth noticing that using only text guidance leads to false facial appearances.
}

\begin{figure}[htbp]
    \vspace{.75mm}
    \centering
    \includegraphics[width=1.00 \linewidth]{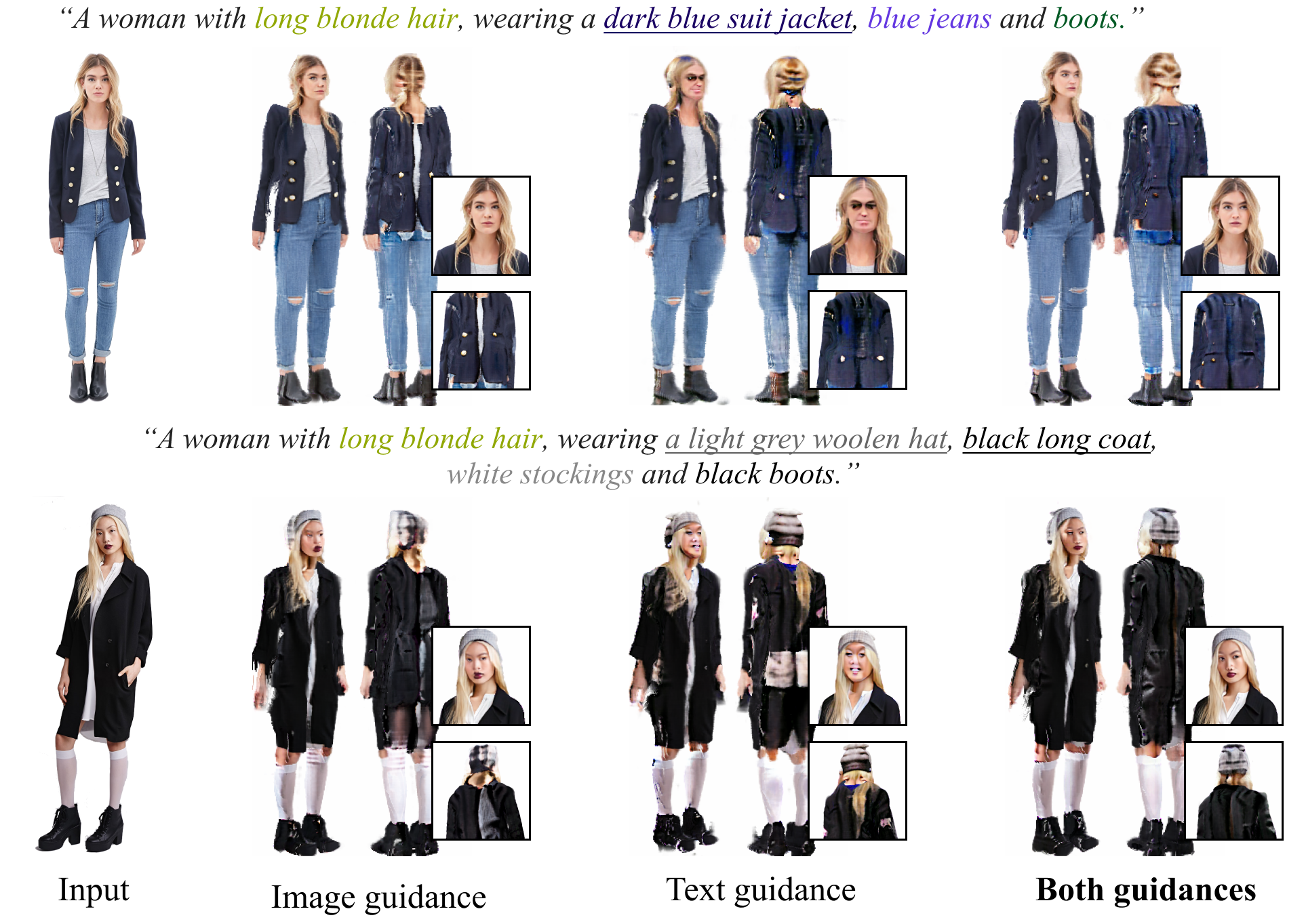}
    \vspace{-6mm}
    \captionof{figure}{\footnotesize\textGuidedCaption}
    \label{fig:text-guided}
    \vspace{-2.5mm}
\end{figure}

\subsubsection{SMPL-based geometric prior}

\indent By incorporating off-the-shelf pose estimation models~\cite{li2022cliff}, we can obtain information about approximate body shapes from SMPL. Our method utilizes this human-specific prior as a geometric clue for 3D human reconstruction and animation by introducing an SMPL-based NeRF initialization and a soft geometry constraint in training.

\nospaceheading{SMPL-based NeRF initialization.} It is difficult for a NeRF model to recover the exact body shape because of occlusions and depth ambiguity. Thus directly optimizing a NeRF with a single image is likely to result in representation degeneration. Inspired by AvatarCLIP~\cite{hongAvatarCLIPZeroShotTextDriven2022}, we initialize our HumanNeRF implicit representation by SMPL meshes renderings. More specifically, we use detected body shape parameters along with pose parameters of the target motion sequence to construct corresponding animated SMPL meshes. Then the multi-view renderings of the meshes are used as pseudo ground truth for initialization. 

Given the estimated parameterized body shape $\beta$ and target motion sequence $\Theta_t=\{\theta_t^i\}_{i=1}^L$, we render image views $\{{I_\textrm{SMPL}}_i^{(j)}\}_{i=1,j=1}^{L, m}$ with pre-defined $m$-view camera poses $\boldsymbol E_s = \{\mathbf{e}_s^i\}_{i=1}^m$ and template meshes generated by SMPL model ${M_i = M_{\textrm{SMPL}}(\beta, \theta_t^i; \Phi)}_{i=1}^{L}$. We also use a template texture to avoid body part occlusion ambiguity. We initialize HumanNeRF with a multi-view setup of its training process. Each iteration samples an image view ${I_\textrm{SMPL}}_i^{(j)}$ for training with a reconstruction loss on the result.

\nospaceheading{Soft geometry constraint.} In the initialization stage, We utilize the human body geometry with SMPL meshes to help the model render approximate shapes for the target character in specific poses.
However, we empirically find that optimizing the model with only the semantic loss may lead to degenerated results, including inconsistent rendered poses and missing body parts, despite the similarity of the CLIP embedding across various views and poses.

For this issue, we introduce a soft geometry constraint based on the assumption that the estimated SMPL meshes are close to the geometry of the naked body of the target character. Therefore, the estimated SMPL meshes should be covered by the actual shape of the clothed character. This loss function can be viewed as a masked version of the silhouette loss in \cite{zhangPerceiving3DHumanObject2020}, which consists of an MSE loss and a one-way Chamfer distance loss for the silhouette boundary. We only compute this loss for the rendered alpha pixels that are covered by the SMPL silhouette. Given the SMPL silhouette mask $S$ and the rendered alpha map $A$, we compute the following loss function: 
\begin{equation}
\begin{split}
\mathcal{L}_\textrm{sil}\! =  \!\sum_{p \in S}\!\normmse{A(p) \!-\! S(p)} \!+ \min_{\hat{p}\in \textrm{Edge}(S)}A(p)\normabs{p\!-\!\hat{p}}, 
\end{split}
\end{equation}
where $\circ$ denotes the element-wise product, $\textrm{Edge}(S)$ computes the edge of mask $S$, $A(p)$ is the pixel value of $A$ at $p$, and $S(p)$ is the pixel-wise mask of $S$ at $p$. This constraint maintains the character's body structure in target motion during training. Also, it allows the creation of detailed outside geometries that better match the target avatar.

\subsubsection{Hybrid sampling strategy with appearance prior} 

Only enforcing global semantic consistency among novel views and poses can possibly lead to unrealistic artifacts on body parts. To tackle this issue, we proposed a body-part-aware sampling to refine body-part details and a rotation-aware sampling to better recover heavily occluded views. 

\nospaceheading{Body-part aware sampling.} To improve the quality of synthesized details and overcome the resolution constraint of pre-trained CLIP, SinNeRF~\cite{xuSinNeRFTrainingNeural2022a} proposes using semantic feature loss between extracted features of randomly sampled local patches rather than complete global views. However, in human-specific rendering tasks, appearance and semantic features vary significantly across different body parts. To address this issue, we introduce a body-part-aware patch sampling strategy to synthesize the well-aligned visual details of the human body. 

For each sampled view $V_\textrm{train}=(\theta_\textrm{train}, \mathbf{e}_\textrm{train})$, our method randomly selects a body part $p$ ,including the whole body, to refine. The rendered segmentation of SMPL can determine the corresponding region, $S^p_\textrm{SMPL}(\theta_\textrm{train}, \mathbf{e}_\textrm{train})$, explicitly defined by groupings of SMPL meshes. Accordingly, we adjust the training camera to render a local patch $V^p_\textrm{train}=(\theta_\textrm{train}, \mathbf{e}^p_\textrm{train})$ for this body part. We can also crop
a corresponding reference patch $V^p_s$ from the input image by the SMPL segmentation $S^p_\textrm{SMPL}(\theta_s, \mathbf{e}_s)$. Similarly, we can also render the patch $V^p_{\textrm{ref}}=(\theta_\textrm{train}, \mathbf{e}^p_\textrm{ref})$ when a neighboring view $\mathbf{e}^p_\textrm{ref}$ is sampled as the reference.

\nospaceheading{Rotation-aware sampling for occluded views.} As discussed in \cite{jain2022zero, hongAvatarCLIPZeroShotTextDriven2022}, using a global semantic loss for 3D generation can result in multi-faced appearances on different sides of the object, which are against realism for 3D avatars. To address this issue, we propose an orientation-aware sampling to recover heavily occluded regions.

Specifically, for a sampled pose $\theta_t^i\in\mathbf{\Theta_t}$, we calculate the body orientations (relative to the input image) $\{\psi(\mathbf{e}_s^j)\}_{j=1}^m$ on the horizontal plane of defined camera $\{\mathbf{e}_s^j\}_{j=1}^m$ and divide the cameras into pre-defined ranges of front cameras $\mathbf{E}_\textrm{front}$, side cameras $\mathbf{E}_\textrm{side}$, and rear cameras $\mathbf{E}_\textrm{rear}$ according to $\{\psi_i^j\}_{j=1}^m$. Since body regions of rear views are heavily occluded in the input image, when a rear camera $\mathbf{e}_\textrm{train}\in\mathbf{E}_\textrm{rear}$ is sampled for training, we use the nearest camera $\mathbf{e}_\textrm{ref}\in\mathbf{E}_\textrm{side}$ to render a reference view $V_\textrm{ref}=(\theta_\textrm{train}, \mathbf{e}^p_\textrm{ref})$ instead of the input view $V_s$. Additionally, since the head region of the avatar is more susceptible to multi-faced artifacts, we also use $\mathbf{e}_\textrm{ref}\in\mathbf{E}_\textrm{front}$ for $\mathbf{e}_\textrm{train}\in\mathbf{E}_\textrm{side}$. Such a strategy infers the appearance of totally occluded views with partially visible views, based on the assumption of visual continuity.

\subsubsection{Overall loss function}

ELICIT constructs animatable avatars through a two-stage optimization process. In the SMPL-based initialization stage, we optimize the model using only the reconstruction loss in Eq. (4). In the one-shot training stage, based on the input image and target motion, we optimize the model using the overall loss function consisting of $L_\mathrm{recon}$, $L_\mathrm{CLIP}$, and $L_\mathrm{sil}$. The detailed loss function is defined as follows:
\begin{equation}
\mathcal{L} =
\begin{cases}
\mathcal{L}_\mathrm{recon}, &\mathrm{if}~ V_\mathrm{train}=V_s \\
\lambda_\mathrm{CLIP} \mathcal{L}_\mathrm{CLIP} + \lambda_\mathrm{sil}\mathcal{L}_\mathrm{sil}, &\mathrm{otherwise}
\end{cases}
\end{equation}
where $\lambda_\mathrm{CLIP}$, $\lambda_\mathrm{sil}$ are hyperparameters for different losses. See \supmat for more details of the optimization.

\section{Experiments and Results} \label{sec:experiments}
\newcommand{\compareLargeCaption}{
\textbf{Overall comparison.} Compared with state-of-the-art NeRF based methods\cite{pengNeuralBodyImplicit2021, kwonNeuralHumanPerformer2021, pengAnimatableNeuralRadiance2021a} on novel view synthesis and novel pose synthesis, \mn generates human 3D renderings with more consistent appearance and realistic details from a single image. We adjust the exposure for better visualization. 
}

\begin{figure*}[htbp]
    \vspace{-4mm}
    \centering
    \includegraphics[width=1.0 \linewidth]{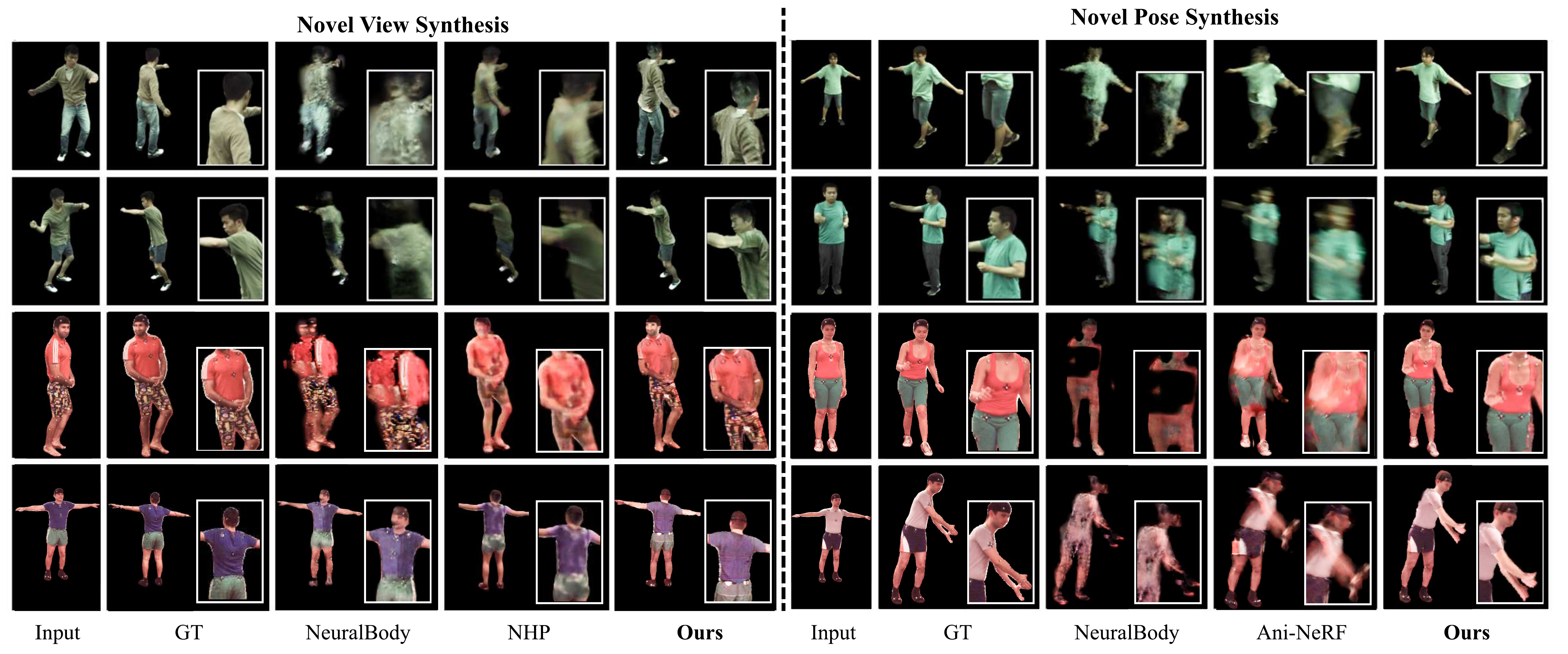}
    \vspace{-6mm}
    \captionof{figure}{\footnotesize\compareLargeCaption}
    \label{fig:compare-large}
    \vspace{-2mm}
\end{figure*}

\subsection{Datasets}
\label{sec:experimentsdatasetes}

We conducted evaluations on two multi-view human video datasets: ZJU-MoCap~\cite{pengNeuralBodyImplicit2021} and Human3.6M~\cite{ionescu2013human36m}, as well as a 2D human image dataset, DeepFashion~\cite{liuLQWTcvpr16DeepFashion}. We selected all nine subjects from ZJU-MoCap and the "Posing" video of all seven subjects from Human3.6M to evaluate free-view animation. To obtain input images, we sampled frames from the first camera of ZJU-MoCap and the third camera of Human3.6M, along with annotated SMPL parameters, camera matrices, and segmentation masks. We applied the annotated motion sequence of each video clip for animation. Additionally, we used high-resolution full-body photos from DeepFashion~\cite{liuLQWTcvpr16DeepFashion} to evaluate our model's performance on human avatars with various clothing styles.

\subsection{Comparison to Existing Methods}

To the best of our knowledge, NeRF-based human-specific novel view synthesis can be classified into per-subject optimization methods and generalizable methods. We selected three state-of-the-art methods as baselines: Neural Body~\cite{pengNeuralBodyImplicit2021} (NB) and Animatable NeRF~\cite{pengAnimatableNeuralRadiance2021a} (Ani-NeRF) from per-subject optimization methods, and Neural Human Performer~\cite{kwonNeuralHumanPerformer2021} (NHP) from generalizable methods. All three methods employ SMPL-based human body priors, with NB and Ani-NeRF supporting novel pose synthesis for animation. For a fair comparison, we adapted these baselines to take single-image inputs. We compared the performance of these methods in two different task settings: novel view synthesis for free-view rendering and novel pose synthesis for character animation.

\nospaceheading{Metrics.} As in previous works~\cite{pengNeuralBodyImplicit2021, kwonNeuralHumanPerformer2021, liuNeuralActorNeural2021a}, we evaluated our results using two standard metrics: peak signal-to-noise ratio (PSNR) and structural similarity index (SSIM). To account for perceptual similarity, we also calculated the Learned Perceptual Image Patch Similarity (LPIPS) metric~\cite{zhang2018unreasonable}, which has been used in recent NeRF-based human rendering works~\cite{wengHumanNeRFFreeViewpointRendering2022, zhaoHumanNeRFEfficientlyGenerated2022}. We followed the evaluation protocol of \cite{pengNeuralBodyImplicit2021} and calculated the metrics only on the bounding box region, rather than the entire image.

\nospaceheading{Comparison on novel view synthesis.} We evaluated the performance of our method and the baselines on the task of novel view synthesis using two multi-view human video datasets, ZJU-MoCap and Human3.6M. We uniformly sampled 10 frames from each subject video and evaluated the results on all available camera views, except for the input view. For per-subject optimization methods NB~\cite{pengNeuralBodyImplicit2021} and Ani-NeRF~\cite{pengAnimatableNeuralRadiance2021a}, we optimized one model for each frame. For the generalizable method NHP~\cite{kwonNeuralHumanPerformer2021}, we sampled three subjects from each dataset as the testing set $S_{test}$ and pre-trained the model only on the remaining subjects of each dataset to ensure a fair comparison.

\begin{table}[htbp]
\centering
\scriptsize
\setlength{\tabcolsep}{3.3pt}
\renewcommand{\arraystretch}{1.15}
\vspace{-2mm}
\begin{tabular}{cccccccc}
&               & \multicolumn{3}{c}{ZJU-MoCAP} & \multicolumn{3}{c}{Human 3.6m} \\
Subjects & Methods                   & PSNR$\uparrow$ & SSIM$\uparrow$ & LPIPS$\downarrow$ & PSNR$\uparrow$ & SSIM$\uparrow$ & LPIPS $\downarrow$\\
\shline
\multirow{2}{*}{$S_{all}$} & NB\cite{pengNeuralBodyImplicit2021} & 20.2          & 0.811 & 0.235 & 20.0     & 0.752    & 0.269    \\
& {\cellcolor{Gray}}ELICIT  & {\cellcolor{Gray}}\textbf{21.9} & {\cellcolor{Gray}}\textbf{0.872} & {\cellcolor{Gray}}\textbf{0.123} & {\cellcolor{Gray}}\textbf{21.5} & {\cellcolor{Gray}}\textbf{0.824} & {\cellcolor{Gray}}\textbf{0.143} \\
\hline
\multirow{3}{*}{$S_{test}$} & NB\cite{pengNeuralBodyImplicit2021} & 19.0          & 0.813 & 0.229 & 20.4     & 0.752    & 0.269    \\
& NHP\cite{kwonNeuralHumanPerformer2021}&21.0 & 0.869 & 0.175 & 21.7     & 0.825    & 0.175    \\
& {\cellcolor{Gray}}ELICIT & {\cellcolor{Gray}}\textbf{21.4} & {\cellcolor{Gray}}\textbf{0.886} & {\cellcolor{Gray}}\textbf{0.118} & {\cellcolor{Gray}}\textbf{21.8} & {\cellcolor{Gray}}\textbf{0.829} & {\cellcolor{Gray}}\textbf{0.146} \\
\end{tabular}
\vspace{-2mm}
\caption{\footnotesize\label{table:compare-nvs} Quantitative comparison of novel view synthesis on ZJU-MoCap and Human3.6M in PSNR, SSIM~(higher is better) and LPIPS~(lower is better). \mn outperforms NB and NHP on all metrics.}
\vspace{-3mm}
\end{table}

As shown in Table \ref{table:compare-nvs}, our \mn outperformed the baselines in terms of PSNR, SSIM and LPIPS on both datasets. Notably, our method's superior performance on the SSIM and LPIPS metrics highlights its advantage in producing perceptually high-quality rendering results. Overall, these results demonstrate the effectiveness of our method in synthesizing high-quality novel views of human subjects.

\nospaceheading{Comparison on novel pose synthesis.} For both datasets, we select one front-view image as input for each subject and evaluate the entire video clip synthesized with motion annotations. For Ani-NeRF, we use the pose-dependent displacement field model proposed in \cite{pengAnimatableImplicitNeural2022}, which reports their best results. As shown in \tab{table:compare-nps}, our method also produces high-quality synthesis when generalized to novel poses.

\begin{table}[htbp]
\centering
\scriptsize
\setlength{\tabcolsep}{4pt}
\renewcommand{\arraystretch}{1.15}
\vspace{-1mm}
\begin{tabular}{ccccccc}
\multirow{2}{*}{Method}               & \multicolumn{3}{c}{ZJU-MoCAP} & \multicolumn{3}{c}{Human 3.6m} \\       & PSNR$\uparrow$ & SSIM$\uparrow$ & LPIPS$\downarrow$ & PSNR$\uparrow$ & SSIM$\uparrow$ & LPIPS $\downarrow$\\
\shline
NeuralBody\cite{pengNeuralBodyImplicit2021} & 20.2 &	0.784 &	0.282  &	21.5 &	0.799 &	0.264      \\
Ani-NeRF\cite{pengAnimatableNeuralRadiance2021a}& 20.3 &	0.791 &	0.277 &	22.2 &	0.807 &	0.229     \\
{\cellcolor{Gray}}ELICIT &  {\cellcolor{Gray}}\textbf{21.8} &	{\cellcolor{Gray}}\textbf{0.853} &	{\cellcolor{Gray}}\textbf{0.143} &	{\cellcolor{Gray}}\textbf{22.6} &	{\cellcolor{Gray}}\textbf{0.859} &	{\cellcolor{Gray}}\textbf{0.123}  \\
\end{tabular}
\vspace{-2mm}
\caption{\footnotesize\label{table:compare-nps} Quantitative comparison of \textbf{novel pose synthesis} on ZJU-MoCap and Human 3.6M in PSNR, SSIM (higher is better) and LPIPS (lower is better). ELICIT outperforms both NB and Ani-NeRF in all the metrics.}
\vspace{-1mm}

\end{table}

We show sampled novel view and pose synthesis results in \fig{fig:compare-large}. Compared to the latest NeRF-based methods, \mn performs better in rendering realistic visual details and inferring occluded contents of clothed human bodies.

\label{sec:experimentsmetrics}

\newcommand{\qualitativeCaption}{
\textbf{Qualitative results} of PIFu~\cite{saitoPIFuPixelAlignedImplicit2019}, PaMIR~\cite{zheng2021pamir}, PHORHUM~\cite{alldieck2022photorealistic} and ELICIT on DeepFashion~\cite{liuLQWTcvpr16DeepFashion}. ELICIT generates more realistic details in occluded views and generalizes well on challenging body poses.
}

\begin{figure*}[tbp]
    \centering
    \includegraphics[width=1.0 \linewidth]{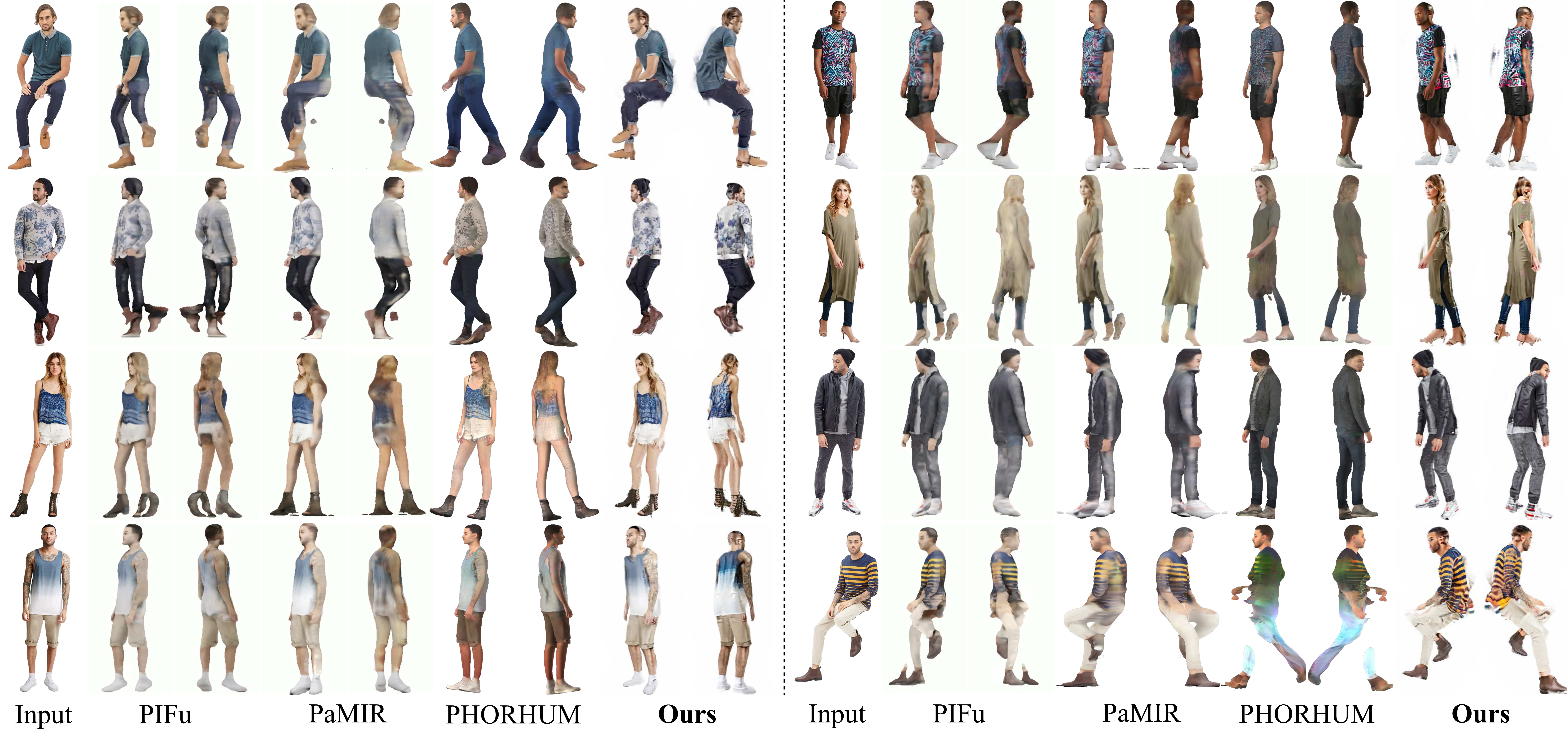}
    \vspace{-8mm}
    \captionof{figure}{\footnotesize\qualitativeCaption}
    \label{fig:qualitative}
    \vspace{-4mm}
\end{figure*}

\subsection{Qualitative Analysis}

Our single-image-based method aims to enable users to create animatable 3D characters from simply available photos of real people. Therefore, in addition to the quantitative evaluation on multi-view human video datasets, we evaluate our approach on 2D human images from DeepFashion~\cite{liuLQWTcvpr16DeepFashion} dataset, with SMPL parameters estimated by off-the-shelf pose estimation models~\cite{li2022cliff, zhang2022pymaf}. Among previous data-driven non-NeRF methods, PIFu~\cite{saitoPIFuPixelAlignedImplicit2019} and PaMIR~\cite{zheng2021pamir} and PHOHRUM~\cite{alldieck2022photorealistic} support both reconstruction of geometry and texture from single image input, which have also shown impressive results on DeepFashion dataset. Here we choose these three methods for qualitative comparison. \Fig{fig:qualitative} illustrates that our training-data-free one-shot method generalizes well on real-world human images and creates rich details for body textures, such as patterns on clothes and shoes, tattoos on the skin, and details of face and hair. While PIFu and PaMIR produce blurry results, limited by the distribution gap between training data and in-the-wild data. 
\subsection{Ablation Studies}
We conduct our ablation studies on introduced model-based priors and select representative subjects from ZJU-MoCap and DeepFashion for comparison. 

\newcommand{\ablationBodypartCaption}{
\textbf{Qualitative results} for the ablation studies of priors used in our method, selected from ZJU-MoCap~\cite{pengNeuralBodyImplicit2021} dataset.
}

\begin{figure}[tbp]
    \centering
    \includegraphics[width=\linewidth]{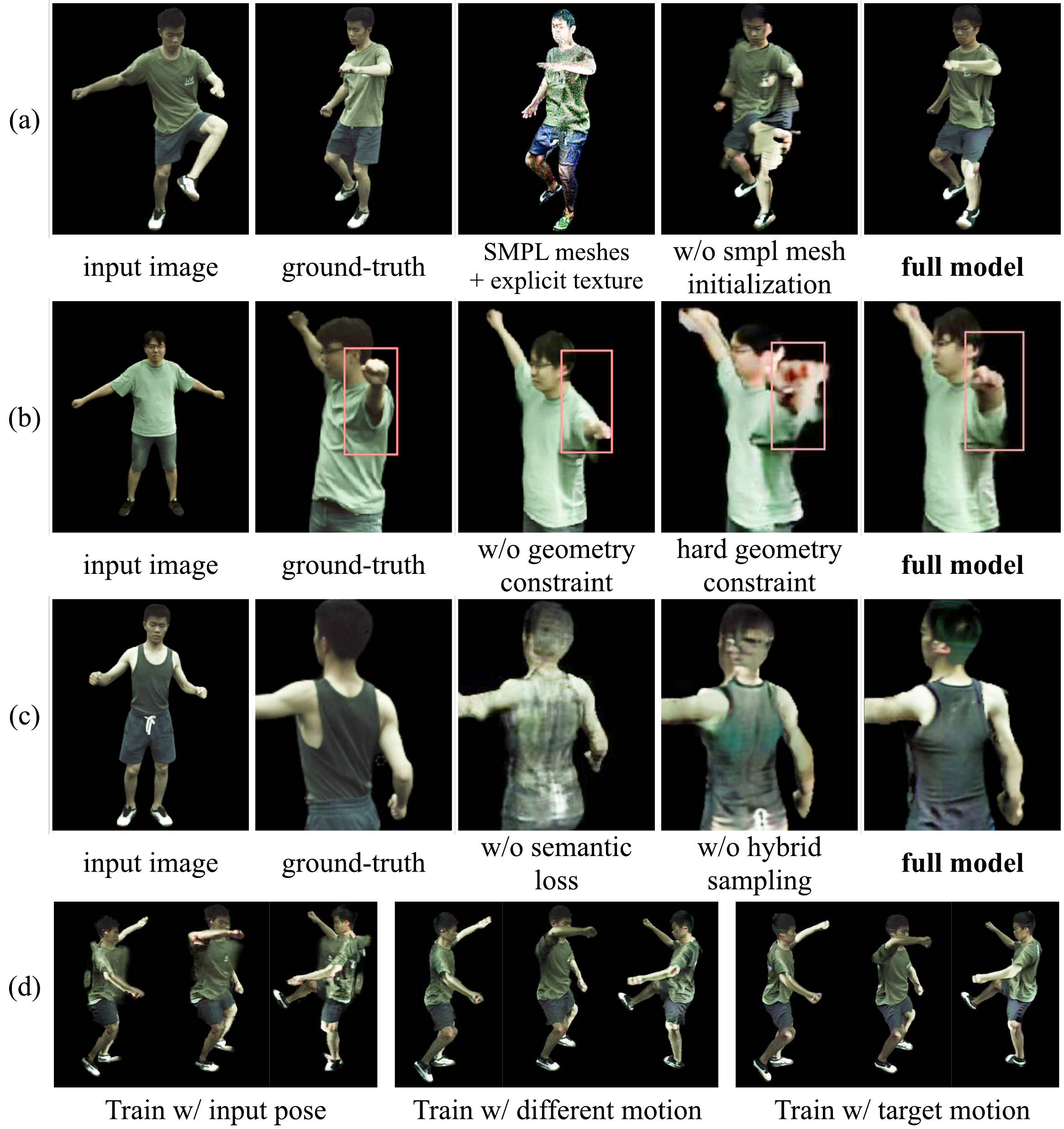}
    \vspace{-6mm}
    \captionof{figure}{\footnotesize\ablationBodypartCaption}
    \label{fig:ablation-bodypart}
    \vspace{-3mm}
\end{figure}

\nospaceheading{Implicit representation.}
We compare our method with a simple baseline of modeling the animatable character explicitly by SMPL meshes, which only optimizes its per-triangle texture parameters during training. Such an explicit model produces noisy textures, and its SMPL-based geometry is also inaccurate compared to the actual human shape. As shown in \tab{table:ablation} and \fig{fig:ablation-bodypart}(a), an implicit representation such as HumanNeRF, which models the character appearance with a spatially continuous function, is necessary for the one-shot learning stage.

\nospaceheading{SMPL mesh initialization.} Initializing our implicit representation with the rendered views of SMPL mesh imparts an approximate human shape and body part semantics at the beginning of the optimization. The significant performance drop in \tab{table:ablation} and \fig{fig:ablation-bodypart}(a) illustrates that this step is necessary for our approach. Only on this basis can semantic loss and geometric constraints guide the completion of detailed geometry and textures.

\nospaceheading{Soft geometric constraint.} As shown in \fig{fig:ablation-bodypart}(b), optimizing the model without geometric constraints may lead to error poses Moreover, in contrast to matching the SMPL geometry directly by a hard constraint of silhouette loss, we only penalize the internal misalignment. This soft constraint allows the implicit model to learn human geometry with clothes and affiliate objects, while the hard one brings in artifacts due to the misalignment of the SMPL shape and the clothed body shape.

\nospaceheading{CLIP-based semantic loss.} As shown in \fig{fig:ablation-bodypart}(c), our semantic loss plays a vital role in generating plausible content for the occluded areas. We also compare the performance of different pre-trained vision models in \supmat. The results indicate that vision models pre-trained with large multi-modal data and
large-capacity models are particularly effective in the semantic loss.
\label{sec:ablation_clip}

\nospaceheading{Sampling strategy.} \Fig{fig:ablation-bodypart}(c) illustrates that certain artifacts from some small areas can significantly affect the overall visual quality, such as texture artifacts on cloth textures and missing hair on the back of the head. Our hybrid sampling strategy helps to generate vivid details and avoid multi-faced artifacts for avatar creation.

\nospaceheading{Training poses.} In \Fig{fig:ablation-bodypart}(d), we show a comparison of animation results of different training poses. The failure of body shape in the input-pose-only training result illustrated the necessity of diverse training poses. While the results of training with different motion sequences show that \mn generalizes well to novel poses in test-time animation.

\begin{table}[htbp]
\vspace{-1mm}
\centering
\footnotesize
\resizebox{0.475\textwidth}{!}{
\begin{tabular}{lccc}
Setting                   & PSNR$\uparrow$ & SSIM$\uparrow$ & LPIPS$\downarrow$ \\ \shline 

SMPL mesh w/ explicit texture & 17.20 & 0.7779 & 0.2116 \\
w/o SMPL mesh initialization & 19.01 & 0.7911 & 0.2122     \\
w/o semantic loss         & 21.46 & 0.8592 & 0.1344     \\
w/o geometric constraint & 21.68 & 0.8633 & 0.1282     \\
hard geometry constraint & 20.58& 0.8288& 0.1664 \\
\multicolumn{1}{l}{w/o hybrid sampling strategy} & \multicolumn{1}{c}{21.46} & \multicolumn{1}{c}{0.8592} & 0.1344  \\
training only w/ input pose & 21.47 & 0.8562 & 0.1516 \\
\textbf{full model}                & \textbf{22.61}  &\textbf{0.8908}  & \textbf{0.1115}         \\
\end{tabular}
}
  \vspace{-2mm}
\caption{\label{table:ablation}\footnotesize Ablation study on subjects \{313, 377, 392\} of ZJU-MoCap. }
    \vspace{-4mm}
\end{table}
\label{sec:experimentablation}
\section{Discussion on Limitations}

While our reconstruction results are generally promising, there remain certain instances of failure. This section provides a comprehensive analysis regarding the limitations of \mn and discusses some potential future directions.

\newcommand{\limitationsCaption}{
We show failure cases of (a) mirrored appearance, (b) geometry artifacts, and (c) errors in texture re-projection for limitation analysis.
}

\begin{figure}[htbp]
    \vspace{-1mm}
    \centering
    \includegraphics[width=1.0 \linewidth]{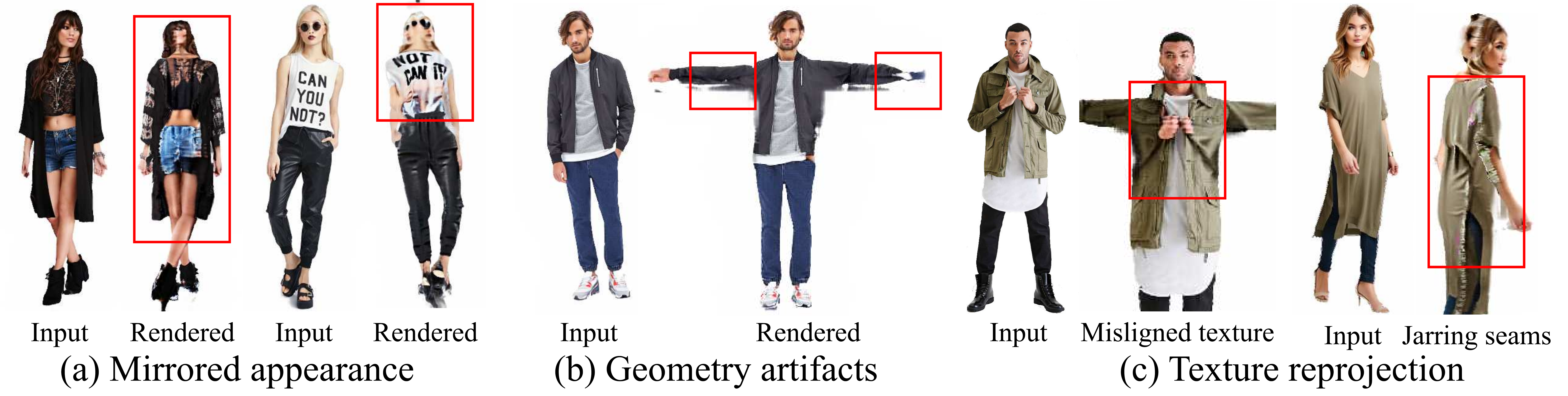}
    \vspace{-6.5mm}
    \captionof{figure}{\footnotesize\limitationsCaption}
    \label{fig:limitations}
    \vspace{-.5mm}
\end{figure}

\qheading{Mirrored appearance}
Although \mn can successfully recover the back-side appearance in many cases with the help of $\mathcal{L}_\mathrm{CLIP}$ and the hybrid sampling strategy, the problem of mirrored appearance still happens sometimes. As shown in \Fig{fig:limitations}(a), since \mn cannot separate the semantic information of different attributes in $\mathcal{L}_\mathrm{CLIP}$, it fails to recover complex garment layering and pattern. The presence of intricate facial attributes can also result in mirrored faces.
In \Fig{fig:text-guided}, we have shown a possible improvement to $\mathcal{L}_\mathrm{CLIP}$ with text guidance. We believe that enhancements with richer semantic information (e.g., human parsing segmentation~\cite{ATR} and view-aware text guidance~\cite{hongAvatarCLIPZeroShotTextDriven2022}), and integration of text-to-image generative models~\cite{saharia2022photorealistic, poole2022dreamfusion,qiu2023controlling} can further improve the quality of back-side appearance.

\qheading{Limited geometry quality}
Apart from an SMPL-based initialization and a soft geometry constraint, \mn has no direct supervision of the clothed body geometry and relies on $\mathcal{L}_\mathrm{CLIP}$ to create geometry details indirectly. As shown in \Fig{fig:limitations}(b), the artifacts in the self-contact body parts and hands show $\mathcal{L}_\mathrm{CLIP}$ has limited ability in modeling geometric details. 
In future work, to alleviate this problem, we can introduce additional supervision of the geometry, including surface regularization, estimated surface normal and depth, and accurate estimation of face geometry and hand geometry, thus enabling expressive animation from some motion generation methods, \eg, TalkSHOW~\cite{yi2023generating}.

\qheading{Texture re-projection}
We use a strong constraint of $\mathcal{L}_\mathrm{recon}$ to re-project the input view texture. However, as shown in \cref{fig:limitations}(d) some texture could be reprojected onto the wrong body parts due to the misalignment between the recovered body shape and the actual geometry, and a strong loss weight of $\mathcal{L}_\mathrm{recon}$ at the edge of the input could lead to jarring seams. Improving geometry alignment, disentangling and balancing $\mathcal{L}_\mathrm{recon}$ of different body parts could be potential solutions for this problem.

\vspace{-0.5mm}
\section{Concluding Remarks} \label{sec:conclusion}
\vspace{-0.6mm}
We introduce ELICIT, a novel method to construct an animatable implicit representation from a single image input and generate a free-view video of the character in the target motion. Two model-based priors drive the one-shot optimization of ELICIT: the visual-model-based visual semantic prior and the SMPL-based human body prior, which enables the reconstruction of body geometry and the inference of full body clothing. We evaluate our methods both qualitatively and quantitatively. We demonstrate our superior performance in single-image settings compared to prior work on novel view and novel pose synthesis, and strong generalizability on real-world human images.

\vspace{1.25mm}
{
\begin{spacing}{0.85}
{\small \noindent\textbf{Acknowledgements}. This work was supported in part by the German Federal Ministry of Education and Research (BMBF): Tübingen AI Center, FKZ: 01IS18039B, in part by The National Nature Science Foundation of China (Nos: 62273301, 62273302, 62273303, 62036009, 61936006), in part by the Key R\&D Program of Zhejiang Province, China (2023C01135), and in part by Yongjiang Talent Introduction Programme (No: 2022A-240-G).}
\end{spacing}
}

\pagebreak

{\small
\balance
\bibliographystyle{ieee_fullname}
\bibliography{00_BIB}
}

\clearpage
\newpage

\begin{appendices} 
\label{appendices}
\section{Implementation Details} \label{sec:implDetails}

In this section, we provide important implementation details for our experiments. 
\ificcvfinal 
We also publicly release our experiment code, results, and model checkpoints at \url{https://huangyangyi.github.io/ELICIT} for research purposes.
\else 
We also include our experiment code, results, and model checkpoints in the supplementary materials for reproducibility, and the code will be publicly available for research purposes.
\fi
\subsection{Optimization}
 In this section, we provide details about the two-stage optimization process of ELICIT.
 For loss weights settings in Eq. (7), we set
 $\lambda_\mathrm{CLIP}=0.1$, $\lambda_\mathrm{sil}=0.01$ are the loss weights.
 We do not use text prompts in our experiments unless specified, for a fair comparison with baseline methods. The initialization stage takes $T_\mathrm{init}=15,000$ iterations of optimization, while the one-shot training stage takes $T_\mathrm{train}=20,000$ iterations. The entire training process for each subject takes approximately 5 hours on 4 NVIDIA Tesla V100 GPUs. We follow the hyper-parameter settings of the HumanNeRF\cite{wengHumanNeRFFreeViewpointRendering2022} code for the optimizer, learning rate, and ray sampling configurations. Specifically, we only train $T_\mathrm{train}=5,000$ for quantitative comparison on novel view synthesis in Tab. 2. 

\subsection{Details of hybrid sampling strategy}
In this section, we provide a detailed description of our hybrid sampling strategy, which combines body-part-aware sampling and rotation-aware sampling in one-shot training.

For each iteration, we randomly decide whether to sample a novel view from $\{(\theta_i, \mathbf{e}_j)\}_{i=1,j=1}^{L,M}$ or the input view $V_s=(\theta_s, \mathbf{e}_s)$ with a probability of $p_\mathrm{novel}\!=\!0.5$. If $V_\mathrm{train}\!=\!V_s$, we follow HumanNeRF to sample a pair of patches for reconstruction. Otherwise, we randomly select a body part $k$ (including the whole body) with weighted probability$\{p_\mathrm{part}^k\}_{k=1}^K$, and sample a training patch $V_\mathrm{train}^k$ which is decided by the bounding box of SMPL rendered body-part segmentation $S_\mathrm{SMPL}^k(V_\mathrm{train})$. 

After sampling the training patch, we sample the reference patch from the $V_s$ or other views of the same pose $\{(\theta_\mathrm{train}, \mathbf{e}_j)\}_{j=1, j\neq i}^{M}$. The camera views of the current pose are divided into front views, rear views, left views, and right views according to the body rotation angle. We assume that the input image is close to the front view of the character. If a rear view of specific body parts (e.g. head, upper body, or whole body) is sampled as the training view, we randomly sampled nearest views from left views and right views as $V_\mathrm{ref}$. Then we render body-part patch $V_\mathrm{ref}^k$ by our NeRF model as reference. Otherwise, the reference patch will be constructed by the resized patch $V_s^k$ cropped from the input image. We set the size of patches in training to 224$\times$224 for all experiments, the same as the input resolution of the CLIP ViT/L-14 model we use for semantic prior.

\subsection{Detailed configuration of evaluation}
In this section, we provide the detailed configuration of our quantitative comparison on ZJU-MoCap dataset and Human 3.6M dataset.

\subsubsection{Data splitting}

For per-subject optimization methods Animatable NeRF\cite{pengAnimatableNeuralRadiance2021a} (Ani-NeRF) and NeuralBody\cite{pengNeuralBodyImplicit2021} (NB), we use all subjects of ZJU-MoCap data-set (313, 315, 377, 386, 387, 390, 392, 393, 394) and the "Posing" sequences of Human 3.6M dataset (S1, S5, S6, S7, S8, S9, S11). We provide information on the single input frame of each subject to evaluate novel pose synthesis, and the 10 frames of each subject we sampled to evaluate novel view synthesis in our experiment code.

For Neural Human Performer\cite{kwonNeuralHumanPerformer2021} (NHP), since it requires pre-training on subjects from the same dataset, we only evaluated NHP with 3 testing subjects from each dataset: ZJU Mocap (313, 315, 387), Human 3.6M (S8, S9, S11), and use remaining subjects for pre-training.

\subsubsection{Baseline settings}

\nospaceheading{Neural Human Performer}\cite{kwonNeuralHumanPerformer2021}. We modify NHP to take only one input view from the first camera of ZJUMoCAP or the third camera of H36M and train the model with novel view ground truth from all other available cameras. We keep other hyperparameters the same as original paper and trained each model with 1000 epochs.

\nospaceheading{NeuralBody}\cite{pengAnimatableNeuralRadiance2021a}. We train NB models for each input frame by optimizing the model only on the single input image. We set the number of optimization iterations to 50K, which is enough for NB to converge on the input image (total loss $<$ 0.0001). We keep other hyper-parameter the same as original paper.

\nospaceheading{Animatable NeRF}\cite{pengAnimatableNeuralRadiance2021a}. We choose Ani-NeRF with pose-dependent fields (PDF), which presents the best results in the original paper. We also train Ani-NeRF models until convergence, similar to the setting of NB.

\ificcvfinal
\else
\subsubsection{Evaluation Protocals}
As we discuss in the paper, we follow the evaluation protocol of \cite{pengNeuralBodyImplicit2021} and calculated the metrics only on the bounding box region. However, we find that different code bases have different ways to calculate the 3D bounding boxes. For a fair comparison, we use bounding boxes calculated by NHP for novel view synthesis experiments, and boxes from Ani-NeRF for novel pose synthesis experiments.
\fi

\newcommand{\comparePHORHUMCaption}{
\footnotesize
\textbf{Qualitative comparison with PHORHUM \cite{alldieck2022photorealistic}.} To compare with PHORHUM, we render the albedo of its results without light effect. While PHORHUM generates better geometry and textures on occluded areas of clothed human bodies, it fails to recover the human body shape in challenging poses, as shown in the comparison results.
}

\section{Additional Results}\label{sec:extraExp}

\subsection{Comparison with MonoNHR}

To compare our method with MonoNHR\cite{choiMonoNHRMonocularNeural2022}, which reports state-of-the-art results on human-specific novel view synthesis from a single monocular input, we present qualitative results of MonoNHR and ELICIT on the ZJU-MoCAP dataset. As full results from MonoNHR are not available, we use the novel view synthesis results from its  \href{https://www.youtube.com/watch?v=9-hfGf7dRw4}{official qualitative video} and compare them with the same input view on ELICIT.

As shown in Figure \ref{fig:compare-sota}, while MonoNHR can estimate approximate clothed body geometry, it produces blurry contents on the novel views, whereas ELICIT generates more realistic details on human faces, bodies, and clothing.

\newcommand{\compareSOTACaption}{
\footnotesize
\textbf{Qualitative comparison with MonoNHR\cite{choiMonoNHRMonocularNeural2022}.} While MonoNHR produces blurry faces, ELICIT generates realistic facial details, demonstrating the superior performance of our method.
}

\begin{figure}[htbp]
    \centering
    \includegraphics[width=1.00 \linewidth]{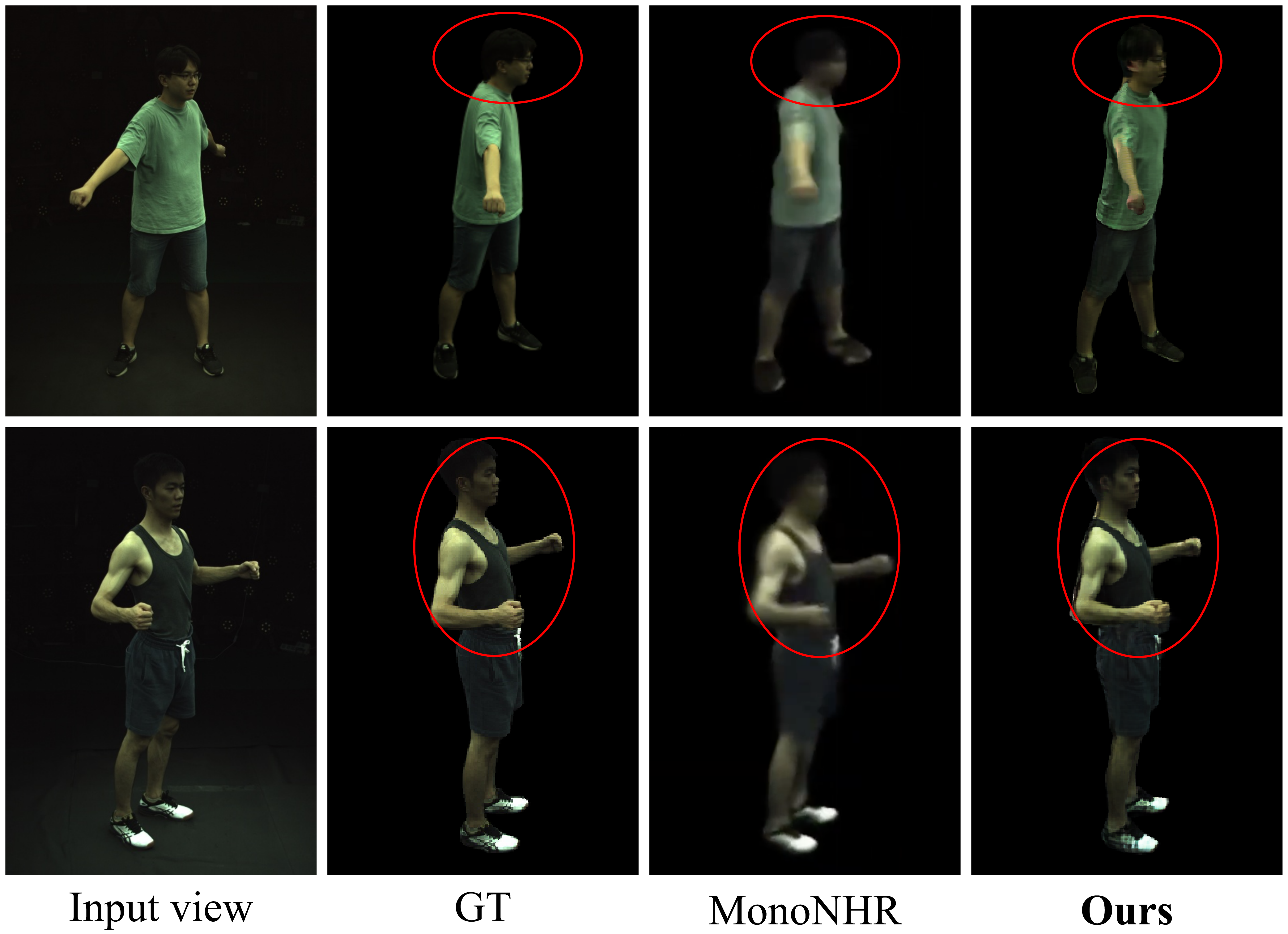}
    \vspace{-6mm}
    \captionof{figure}{\compareSOTACaption}
    \label{fig:compare-sota}
    \vspace{-4mm}
\end{figure}

\subsection{Ablation Study}

\subsubsection{Differerent pretrained visual models}
As discussed in Section 4.4, we also compare the performance of different pre-trained visual models, including an DINO~\cite{caron2021emerging} ViT used by SinNeRF~\cite{xuSinNeRFTrainingNeural2022a}, an ImageNet pretrained ViT/L-14~\cite{dosovitskiyImageWorth16x162021, deng2009imagenet}, an unsupervised pre-trained ViT/L-14 by MAE~\cite{he2022masked}, also a lighter version of CLIP ViT/B-32. As shown in \fig{fig:ablation-semantic}, CLIP ViT/L-14 shows best performance in capturing 3D-aware human body structure and generating vivid visual details, and the two CLIP pre-trained models have a better performance on head structure than Image pre-trained models. This comparison suggests that the rich pre-training data of the CLIP model, as well as the larger model capacity of CLIP ViT/L-14 compared to CLIP ViT/B-32, are key factors contributing to the effectiveness of our semantic loss.

\newcommand{\ablationSemanticCaption}{
Qualitative results for the ablation studies of vision models used for the semantic loss, selected from DeepFashion\cite{liuLQWTcvpr16DeepFashion} dataset. The CLIP ViT/L-14 model we use produce best detailed geometry and textures. 
}

\begin{figure}[thbp]
\vspace{-4mm}
    \centering
    \includegraphics[width=0.9 \linewidth]{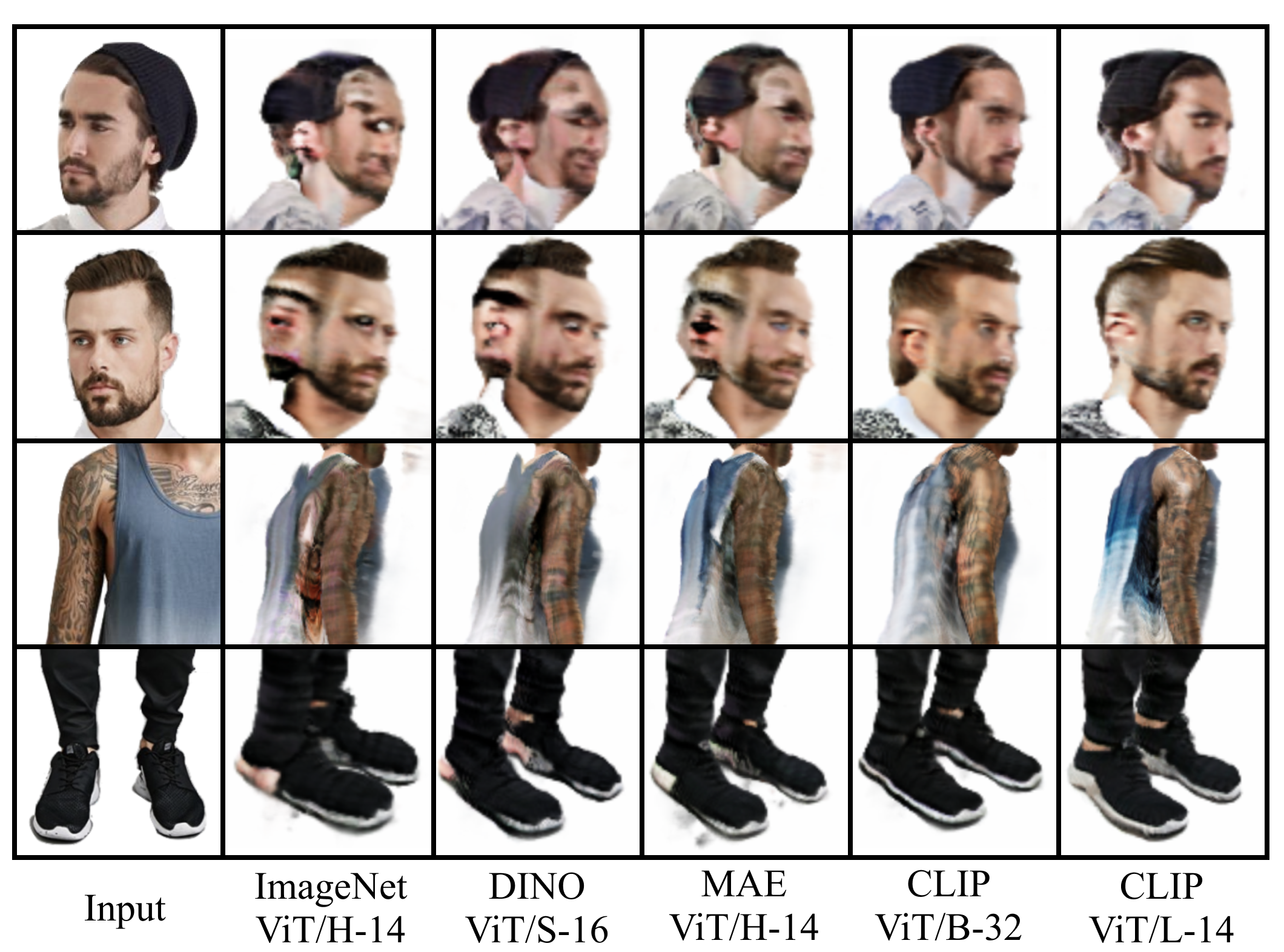}
    \vspace{-4mm}
    \captionof{figure}{\footnotesize\ablationSemanticCaption}
    \label{fig:ablation-semantic}
    \vspace{-4mm}
\end{figure}

\subsubsection{Hybrid sampling strategy}

To thoroughly evaluate the effectiveness of our proposed hybrid sampling strategy, we conducted a detailed ablation study on both body-part-aware sampling and rotation-aware sampling. As shown in \Fig{fig:ablation-sampling}, our results indicate that body-part-aware sampling improves ELICIT's ability to synthesize realistic details on crucial body parts with fine-grained supervision. Additionally, rotation-aware sampling successfully avoids artifacts of mirrored appearance by using neighboring views as a reference to recover heavily occluded body regions.

\newcommand{\ablationSamplingCaption}{
\footnotesize
\textbf{Ablation study of hybrid sampling strategy.} Comparison of training with different sampling strategies: without (a) body-part-aware sampling, without (b) rotation-aware sampling, and full hybrid sampling strategy. The absence of either sampling strategy leads to artifacts, such as mirrored appearance or missing details on important body parts.
}

\begin{figure}[htbp]
    \centering
    \includegraphics[width=1.00 \linewidth]{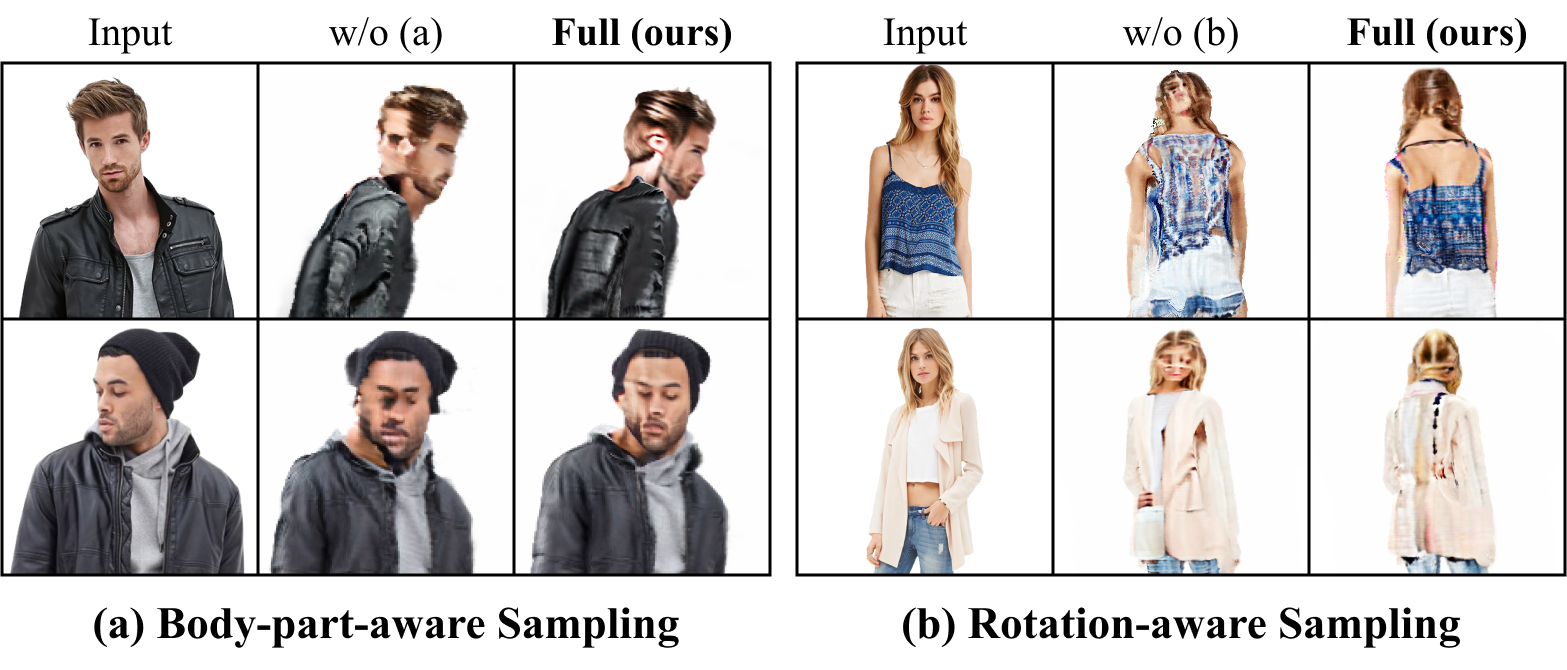}
    \vspace{-6mm}
    \captionof{figure}{\ablationSamplingCaption}
    \label{fig:ablation-sampling}
    \vspace{-4mm}
\end{figure}

\subsubsection{Comparing CLIP loss with perceptual losses}

In our main paper, we compared our CLIP-based semantic loss with various embedding losses that capture high-level semantics. However, since the CLIP loss can also capture low-level visual attributes such as color and texture, we further evaluated its effectiveness by comparing it with two commonly-used perceptual losses: LPIPS\cite{zhang2018unreasonable} and VGG-based perceptual loss\cite{johnson2016perceptual}, in generative and reconstruction tasks. As depicted in \fig{fig:ablation-perceptual}, LPIPS loss and VGG loss only capture a subset of low-level visual features and cannot synthesize 3D-aware appearance with high-fidelity details in occluded areas, unlike CLIP-loss.

\newcommand{\ablationPerceptualCaption}{
\footnotesize
\textbf{Comparison of CLIP loss to other perceptual losses.} LPIPS and VGG-based perceptual losses only capture a subset of low-level visual features, leading to limited performance in synthesizing occluded clothed body appearance compared to CLIP loss.
}
\begin{figure}[htbp]
    \centering
    \includegraphics[width=1.00 \linewidth]{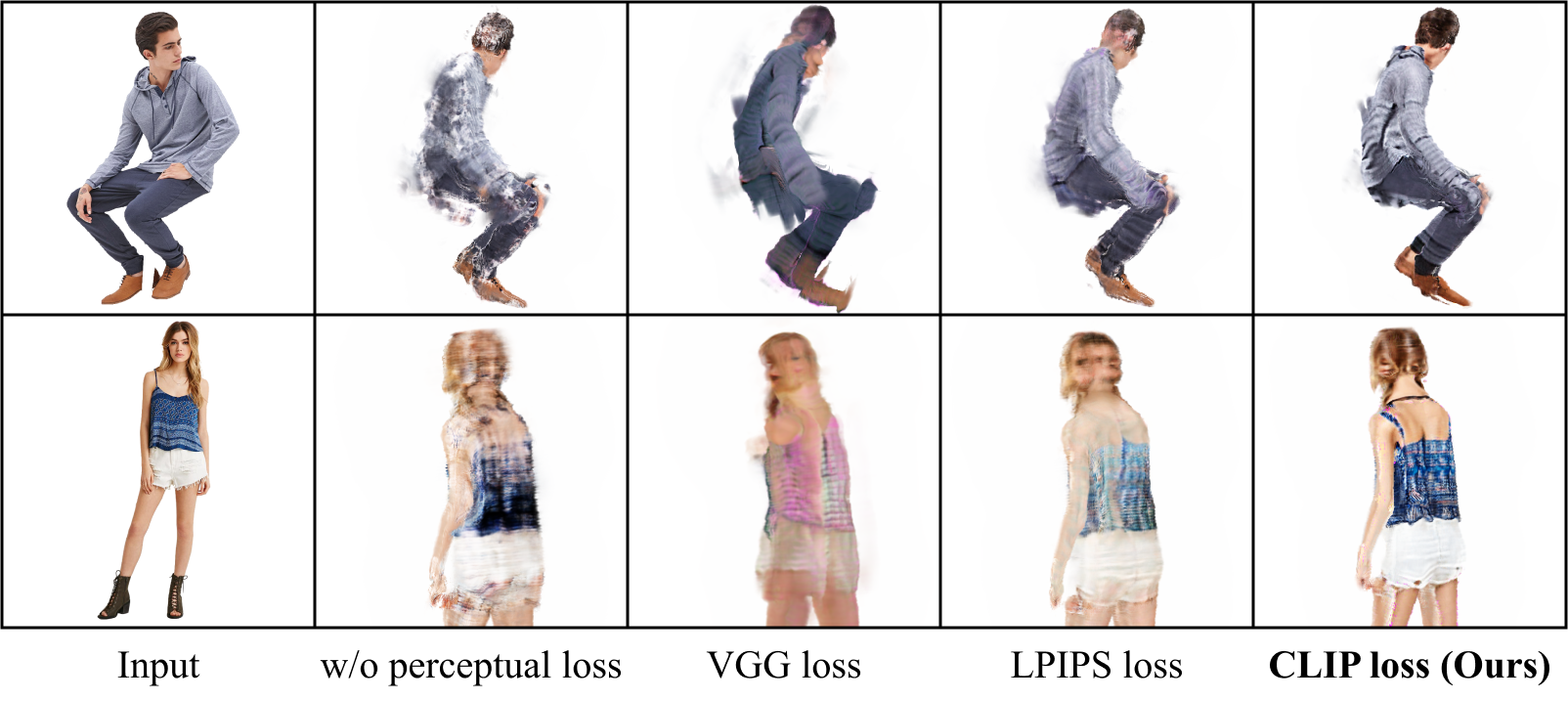}
    \vspace{-8mm}
    \captionof{figure}{\ablationPerceptualCaption}
    \label{fig:ablation-perceptual}
\end{figure}

\subsection{Extensions}

ELICIT proposes a simple and effective pipeline for creating animatable avatars with implicit representation and model-based prior. The pipeline is also extensible for future improvements with different implicit human representations, semantic priors, geometric priors, and input settings. In this section, we introduce several extensions of ELICIT that can inspire future work.

\subsubsection{Alternative human representations}

ELICIT can be trained using various implicit human representations. For example, as shown in \fig{fig:extension-neus}, we replaced the HumanNeRF model used in ELICIT with an SDF-based model from Animatable NeRF\cite{pengAnimatableImplicitNeural2022, pengAnimatableNeuralRadiance2021a}. This alternative representation performed better in surface geometry, while HumanNeRF produced blurry floating artifacts near the body that decreased the rendering quality. Such explorations with different implicit human representations can lead to further improvements in the quality of the synthesized avatars.

\newcommand{\extensionNeusCaption}{
\footnotesize
\textbf{Improved human representation.} The SDF-based model from Ani-NeRF\cite{pengAnimatableImplicitNeural2022} reduces floating artifacts (marked with red rectangles), which are commonly present in our HumanNeRF-based model, leading to better surface geometry.
}
\begin{figure}[htbp]
    \centering
    \includegraphics[width=1.00 \linewidth]{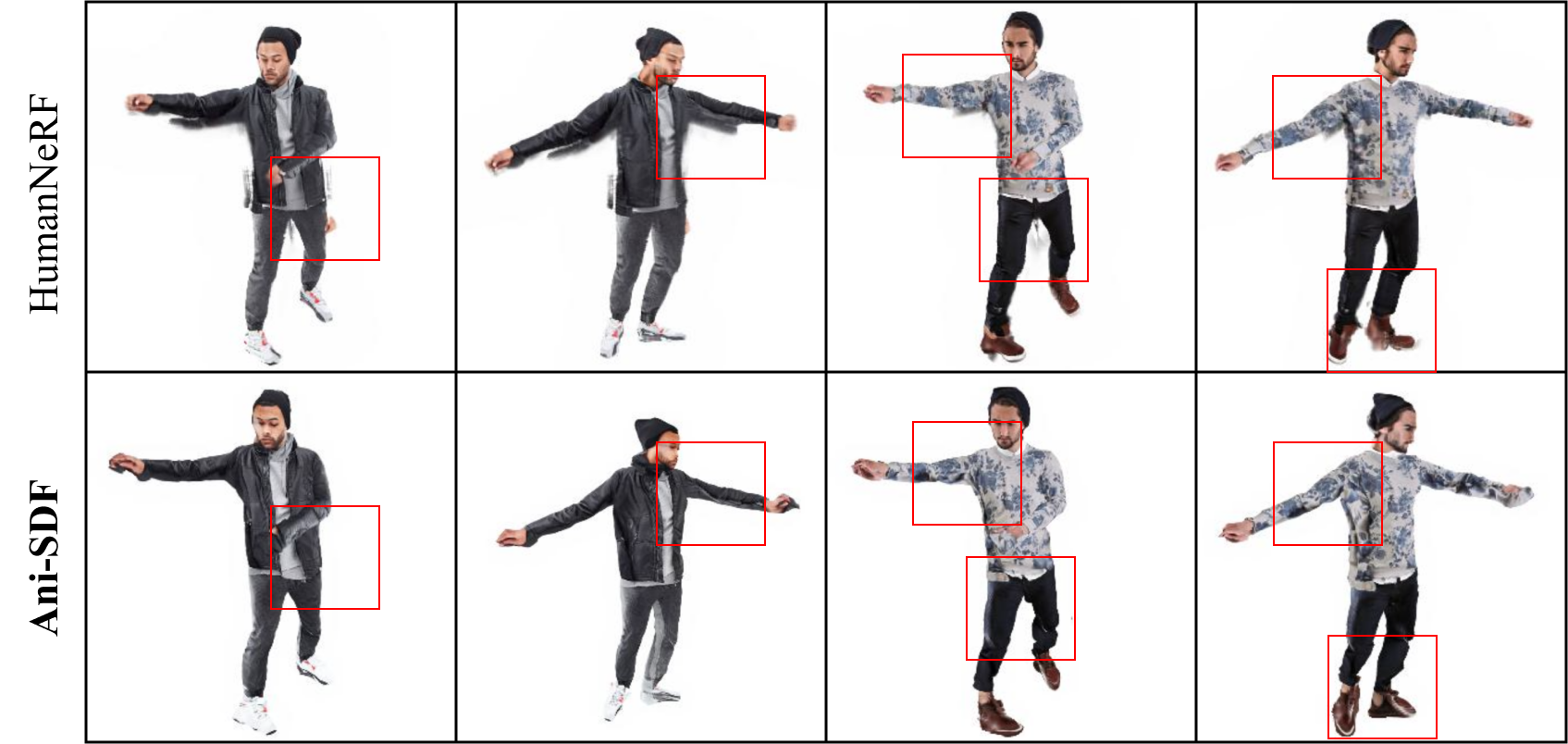}
    \vspace{-6mm}
    \captionof{figure}{\extensionNeusCaption}
    \label{fig:extension-neus}
    \vspace{-4mm}
\end{figure}

\subsubsection{Editing 3D avatars with textual guidance}

As we discussed in our main paper, we can improve the performance of semantic prior by incorporating user text prompts through text-based CLIP guidance and image-based CLIP guidance. In addition, as shown in \fig{fig:extension-text}, ELICIT can generate different text-conditioned appearances using different text prompts, such as manipulating the occluded texture of clothing. These results demonstrate the potential for using ELICIT's pipeline for digital human editing tasks with further improvements.

\newcommand{\extensionTextCaption}{
\footnotesize
\textbf{Generating text-conditioned appearance.} By using different prompts, we can generate various texture patterns in the occluded area of clothing. While the quality of synthesis is limited, it demonstrates the potential of ELICIT for editing 3D avatars.
}
\begin{figure}[htbp]
    \centering
    \includegraphics[width=1.00 \linewidth]{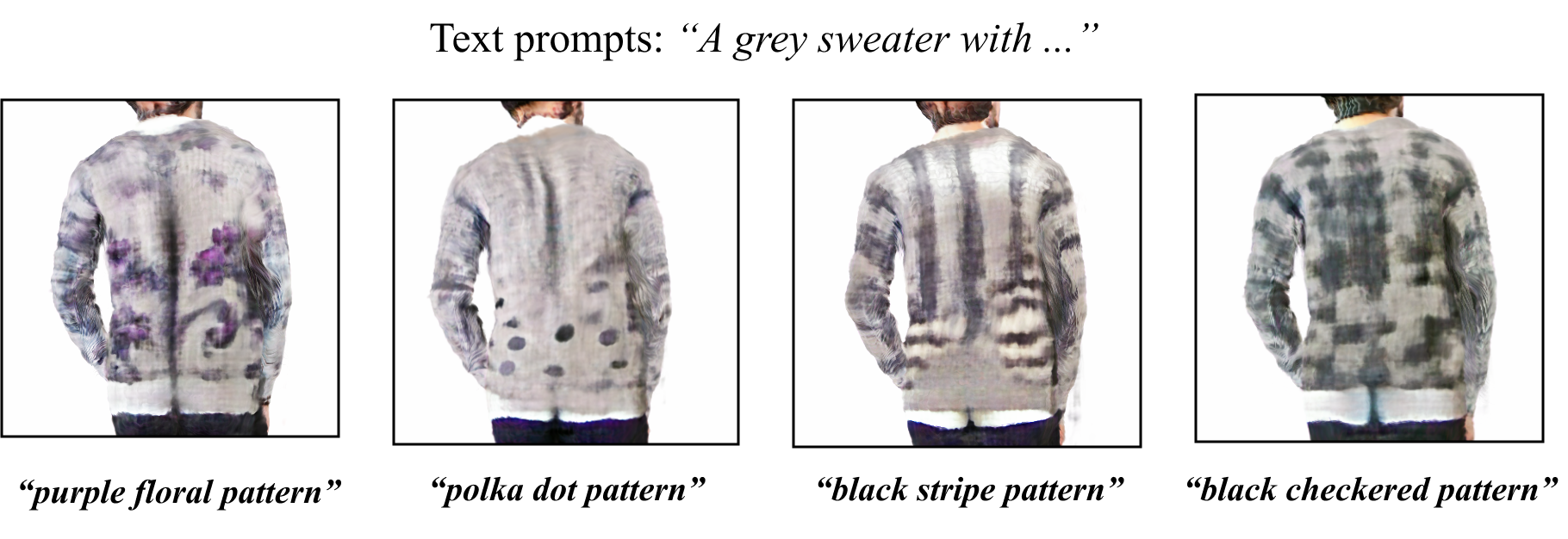}
    \vspace{-6mm}
    \captionof{figure}{\extensionTextCaption}
    \label{fig:extension-text}
    \vspace{-4mm}
\end{figure}

\subsubsection{Utilizing multiple images}

ELICIT can be enhanced by utilizing multiple input images to better recover full-body appearances. It's worth noting that ELICIT can utilize images of different poses without requiring well-aligned pose annotations, by taking one image for reconstruction and using the others as a reference in the CLIP loss. As shown in \fig{fig:extension-multi}, we demonstrate the effectiveness of this approach by incorporating an extra back-side image, resulting in better full-body appearance.

\newcommand{\extensionMultiCaption}{
\footnotesize
\textbf{Utilizing multiple images.} ELICIT can utilize images of different poses as an extra reference to better recover full-body appearance.
}
\begin{figure}[htbp]
    \centering
    \includegraphics[width=1.00 \linewidth]{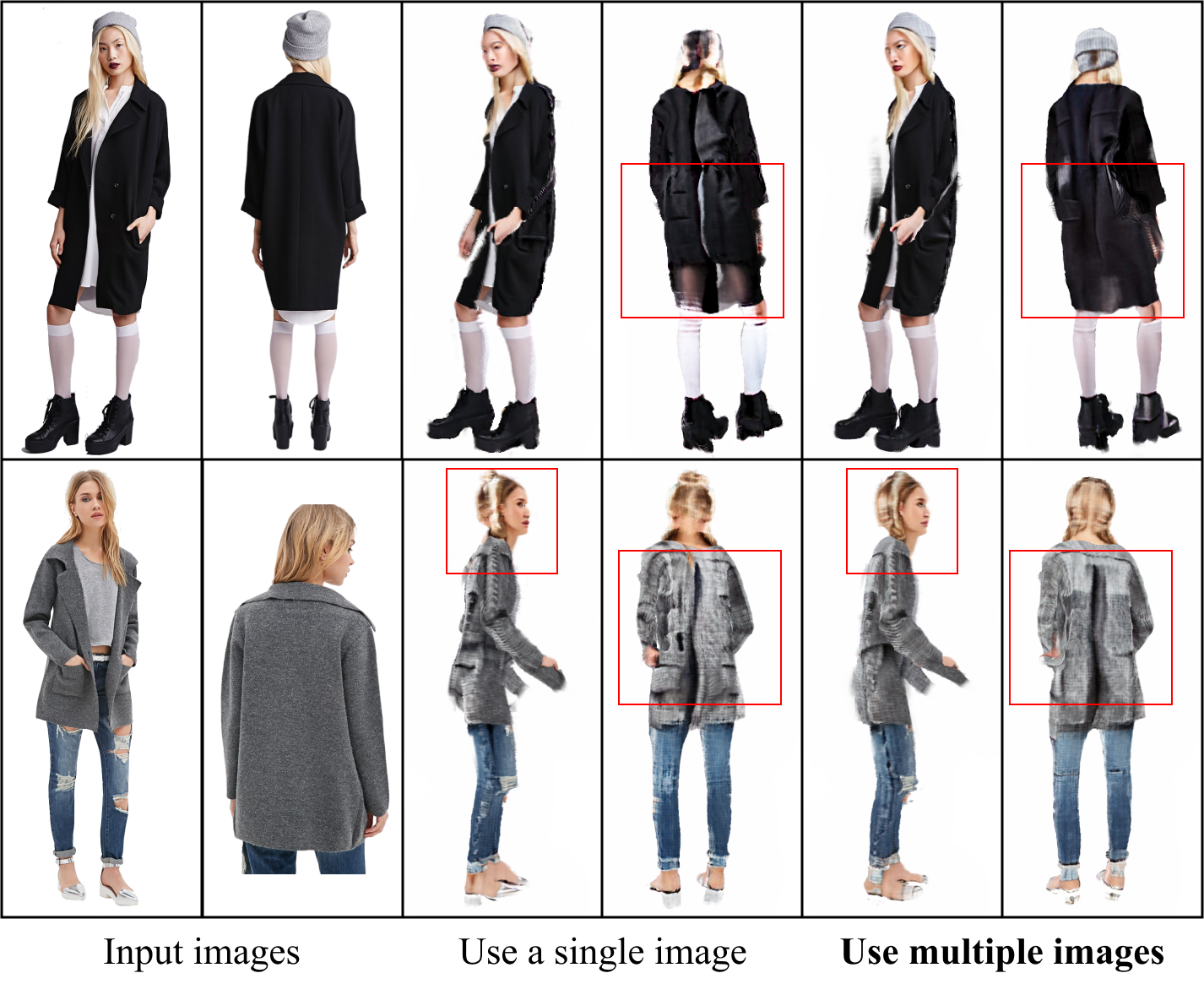}
    \vspace{-6mm}
    \captionof{figure}{\extensionMultiCaption}
    \label{fig:extension-multi}
    \vspace{-4mm}
\end{figure}

\section{Limitations}\label{sec:limitations}

The human body geometry prior utilized by ELICIT requires well-aligned SMPL annotation of body shape and postures. When body parts such as hands and legs are heavily misaligned, artifacts may occur due to the model being initialized incorrectly or failing to sample reference patches for body-part refinement. Furthermore, modeling hand geometry and complex clothing geometry precisely remains a challenge for our method.

Additionally, the computational cost of 5 hours on 4 Tesla V100s per avatar may be prohibitively expensive for certain applications. Future work could focus on developing more efficient human-specific NeRFs that require lower GPU memory, as well as improving the training pipeline to reduce the number of necessary training iterations.

\section{Future Work}\label{sec:futureWorks}

We plan to further explore model-based priors that can potentially improve \mn. For semantic prior, we will investigate the use of image diffusion models \cite{saharia2022photorealistic, nichol2021glide} which have been applied to text-to-3D tasks, as they are promising options for enhancing the appearance details of ELICIT. For geometric prior, we aim to use a more expressive human-body prior with SMPL-X \cite{pavlakos2019expressive} to improve detailed geometry, such as hand shapes. Regarding implicit representation, we are exploring options and improvements with higher efficiency, better surface geometry, and better rendering quality. Additionally, we are working to enhance the versatility of our one-shot training framework to accept different types of inputs (e.g., multiple images, short videos, and images with a text description).
\end{appendices}

\end{document}